\documentclass{article}

\usepackage{graphicx} 
\usepackage{verbatim}
\usepackage[a4paper, margin=0.7in]{geometry}
\usepackage{float}
\usepackage{cite}
\usepackage{mathrsfs}
\usepackage{subcaption}
\usepackage{amsmath}
\usepackage{amssymb}
\usepackage{xcolor}
\usepackage{soul} 
\sethlcolor{yellow} 

\usepackage{tikz}
\usetikzlibrary{positioning,calc,arrows.meta}

\usepackage[colorlinks=true, linkcolor=blue, citecolor=blue, urlcolor=blue]{hyperref}
\usepackage{cleveref}

\title{Dynamics-Informed Deep Learning for Predicting Extreme Events}
\author{Eirini Katsidoniotaki, Themistoklis P. Sapsis
\thanks{Corresponding author: \href{mailto:sapsis@mit.edu}{sapsis@mit.edu},
Tel: (617) 324-7508, Fax: (617) 253-8689%
}\\
Department of Mechanical Engineering,\\
Massachusetts Institute of Technology, \\
77 Massachusetts Ave., Cambridge, MA 02139}
\date{\today}

\begin{document}

\maketitle

\section*{Abstract}

Predicting extreme events in high-dimensional chaotic dynamical systems remains a fundamental challenge, as such events are rare, intermittent, and arise from transient dynamical mechanisms that are difficult to infer from limited observations. Accordingly, real-time forecasting calls for precursors that encode the mechanisms driving extremes, rather than relying solely on statistical associations. We propose a fully data-driven framework for long-lead prediction of extreme events that constructs interpretable, mechanism-aware precursors by explicitly tracking transient instabilities preceding event onset. The approach leverages a reduced-order formulation to compute finite-time Lyapunov exponent (FTLE)–like precursors directly from state snapshots, without requiring knowledge of the governing equations. To avoid the prohibitive computational cost of classical FTLE computation, instability growth is evaluated in an adaptively evolving low-dimensional subspace spanned by Optimal Time-Dependent (OTD) modes, enabling efficient identification of transiently amplifying directions. These precursors are then provided as input to a Transformer-based model, enabling forecast of extreme event observables. We demonstrate the framework on Kolmogorov flow, a canonical model of intermittent turbulence. The results show that explicitly encoding transient instability mechanisms substantially extends practical prediction horizons compared to baseline observable-based approaches.\\

keywords: Extreme events prediction; Dynamical precursors; Time-series forecasting;  
\section{Introduction}

Extreme events are rare yet high-impact episodes, associated with abrupt changes in the state and dynamics of a system, that arise across a variety of natural and engineering systems—including oceanic rogue waves~\cite{dysthe2008oceanic}, extreme weather~\cite{fischer2021increasing}, shocks in power grids~\cite{schafer2018dynamically}, and sudden market drawdowns~\cite{christoffersen1998horizon}, just to mention a few examples. They often lead to severe humanitarian, environmental, and financial consequences, motivating the development of real-time forecasting tools that enable timely mitigation.

To formalize this notion, an extreme event may be defined as a high-amplitude excursion of a system response that markedly exceeds the typical fluctuations generated by the background dynamics~\cite{farazmand2019extreme,sapsis2021statistics}. In time series, such excursions appear as intermittent bursts: short-lived peaks separated by long intervals of typical behavior. In practice, extremes are specified with respect to an observable—a scalar quantity of interest derived from the system state, such as wave crest height in ocean dynamics, 
or portfolio losses in financial markets. An event is deemed extreme when the observable attains values in the far tail of its empirical (or stationary) distribution, and thus occurs with very low probability under nominal conditions.

While this definition provides a clear criterion for identifying extremes, it does not by itself make them predictable: extreme excursions may develop rapidly and often exhibit weak or ambiguous precursory signatures. As a result, real-time prediction hinges on identifying measurable \emph{diagnostic quantities} that characterize the evolving system and exhibit consistent, detectable changes prior to event onset. We refer to such diagnostics as \textbf{precursors} of extremes. For a precursor to be practically useful, it must robustly discriminate impending extremes from typical fluctuations, yielding low false-positive and false-negative rates. In this context, even partial understanding of the dynamics that underlie the extreme events and trigger their formation can guide the discovery and design of reliable precursors that facilitate the data-driven prediction of extremes~\cite{farazmand2019extreme,sapsis2021statistics, barthel2023harnessing}. 


In the context of extreme events associated with special type of dynamics, such as instabilities, special emphasis should be given to mechanism-linked features rather than purely statistical correlations. For example, in unsteady aerodynamic flows, pressure-based precursors—often constructed to isolate event-relevant dynamics (e.g., via spectral or wavelet filtering) and informed by sparse-sensing measurements—have been coupled with sequential time-series models to improve prediction of intermittent force excursions~\cite{rudy2022prediction, barthel2023harnessing, kim2025forecasting}. In turbulent flows with external forcing, mechanism-based precursors of extreme dissipation bursts have been identified by linking burst onset to characteristic nonlinear interactions among Fourier modes and associated depletion of low-wavenumber energy, while the underlying trigger of this energy exchange remains unresolved~\cite{farazmand2017variational, blonigan2019extreme}. A dynamical–statistical approach was proposed for wall-bounded turbulence that identifies precursors of extreme dissipation events by computing an attractor-consistent critical state via finite-time energy-growth optimization and using its alignment with the instantaneous flow as an interpretable early-warning indicator. 
Recent work~\cite{vela2024large} has further shown that dissipation bursts of comparable magnitude can exhibit markedly different predictability depending on their underlying dynamical route, underscoring the need for mechanism-linked, pathway-discriminative precursors. Similarly, in different domains such as medicine, instability-informed scalars, such as data-driven Lyapunov-exponent estimates, have been used in deep-learning models for event detection and forecasting in physiological time series~\cite{varghese2025multifaceted}. 

On the other hand, we have purely data-driven precursors, where predictive signatures of upcoming extremes are learned directly from data and statistical associations~\cite{guth2019machine, chattopadhyay2020analog}. In the same spirit, relevant efforts have increasingly leveraged deep learning methods to predict extreme events directly from time-series data, including recurrent architectures, such as reservoir computing~\cite{pathak2018model, pyragas2020using, doan2021short, ahmed2024prediction} and LSTM-based models~\cite{wan2018data}. To improve performance recent works proposed extreme-event-tailored loss functions~\cite{rudy2023output} and optimal sampling approaches~\cite{blanchard2021output} to emphasize rare events during training. In addition, attention-based sequence models have been proposed to improve performance by explicitly separating “normal” from “extreme” regimes~\cite{Agarwal2025precursors}.  In this purely data-driven paradigm, attribution and explainability methods can be used to extract the input patterns most responsible for a model’s predictions (e.g., for heatwave forecasting), yielding “machine-view” precursors that support knowledge discovery~\cite{wei2025xai4extremes}. While powerful, these precursors are typically association-based, they rely on a large number of extremes for training, whereas dynamics-based precursors are constructed to explicitly quantify the underlying dynamical pathways to extremes.

In this study, we focus on high-dimensional systems in which extremes are generated by \emph{internal transient instabilities} and we build dynamical precursors that encode the responsible dynamics. More specifically, we utilize a geometric viewpoint, where the system evolves primarily on a background attractor where the trajectories lie most of the time. Extreme events occur when the system enters localized regions of this attractor where the dynamics become strongly nonlinear and exhibit finite-time instabilities ~\cite{sapsis2021statistics}. In this case the state is transiently repelled from the main attractor, causing a rapid amplification of the observable that appears as an intermittent burst in the time series~\cite{mohamad2016probabilistic,farazmand2017variational,blonigan2019extreme}. Our goal is to formulate dynamic precursors by encoding the local unstable dynamics associated with these transient events. This is done by relying on the observation that, even in high- or infinite-dimensional systems, the emergence of extreme events is typically dominated by a small number of transiently-positive Lyapunov exponents, i.e. by a small number of effective modes. These modes are hard to capture with traditional spectral decomposition methods —such as dynamic mode decomposition \cite{schmid2010dynamic} and Koopman-mode analysis \cite{mezic2013analysis}— due to their essentially transient character. 


To address this challenge we employ the Optimal Time-Dependent (OTD) framework~\cite{babaee2016minimization}, in a fully data-driven formulation. OTD modes define an adaptive, trajectory-dependent low-dimensional subspace that continuously aligns with the directions of maximal transient growth, i.e. directions associated with the largest finite-time Lyapunov exponents (FTLE), and therefore with the finite-time instabilities that precede extreme excursions~\cite{babaee2017reduced}. The key idea is to discover features that directly track the finite-time instability direction responsible for extreme-event growth and to use those to synthesize a precursor of an upcoming extreme event. This is achieved by computing the FTLE in the reduced-order subspace spanned by the OTD modes. Previous work~\cite{farazmand2016dynamical} relied on purely OTD-based precursors capturing local-in-time transient amplification, whereas our FTLE-based approach encodes instability growth accumulated over a finite horizon. A key remaining challenge is to translate mechanistic, trajectory-based instability diagnostics, in this case FTLE, into reliable forecasts of event occurrence from time-series observations. Recent work has shown that the predictive skill for rare extremes depends not only on the use of dynamically informative, mechanism-linked features but also on their integration with appropriate deep-learning architectures to improve forecasting performance and extend lead times~\cite{asch2022model}. In our case, we address this step by employing a Transformer-based architecture that exploits temporal context to forecast extreme-event occurrence over a prescribed lead time, using the FTLE obtained in earlier times. 

 We demonstrate the method on a prototype high-dimensional turbulent system —the Kolmogorov flow— in which extremes appear as intermittent bursts of the total energy dissipation rate, showing that the proposed framework extends substantially the effective prediction horizon relative to baseline precursors while maintaining computational efficiency, thereby enabling practical early warning in settings where extreme events pose significant risk.

\section{Preliminaries and Background}
\label{sec:Background}

\subsection{Problem Setup}
\label{sec:problem-setup}

We consider a general nonlinear dynamical system whose evolution in the phase space is governed by
\begin{equation}\label{eq:setup_dynsystem}
    \dot{\mathbf{u}} = \mathbf{F}(\mathbf{u}), 
    \qquad \mathbf{u}(t)\in\mathcal{U},
\end{equation}
where $\mathcal{U}$ denotes the state space (e.g., $\mathbb{R}^n$ for a finite-dimensional ODE, or an appropriate Hilbert space for an infinite-dimensional PDE), and $\mathbf{F}:\mathcal{U}\to\mathcal{U}$ is a nonlinear vector field or nonlinear operator governing the system dynamics.
For any initial condition $\mathbf{u}(t_0)=\mathbf{u}_0$, the system state at time $t$ can be written as
\begin{equation}\label{eq:flow_map}
\mathbf{u}(t;\mathbf{u}_0)=\psi_{t_0}^{t}(\mathbf{u}_0),
\end{equation}
where $\psi_{t_0}^{t}:\mathcal{U}\to\mathcal{U}$ is the flow map in the phase space. 
We assume that the long-time dynamics are supported on a chaotic attractor and that the system exhibits rare, intermittent excursions away from typical fluctuations, which we refer to as extreme events.

To characterize the local, finite-time stability properties of the dynamics along a given trajectory, we examine the evolution of infinitesimal perturbations to the state. Let $\boldsymbol{\xi}(t)\in\mathcal{U}$ denote a small perturbation about the reference trajectory $\mathbf{u}(t)$. Linearizing the governing equations about $\mathbf{u}(t)$ yields the variational equation.
Small perturbations superposed on a reference trajectory in a dynamical system can be described as tangent linear evolutions about the trajectory.
\begin{equation}\label{eq:setup_variational}
    \boldsymbol{\dot{\xi}}= \mathbf{L}\,\boldsymbol{\xi},
    \qquad 
    \mathbf{L}(t) := \nabla_{\mathbf{u}}\mathbf{F}(\mathbf{u}),
\end{equation}
where $\mathbf{L}$ is the Jacobian (or Fréchet derivative, in the infinite-dimensional case) of the vector field evaluated along the trajectory. 
Solutions of~\eqref{eq:setup_variational} describe the instantaneous growth or decay of perturbations and provide a trajectory-dependent notion of stability. This is particularly relevant in chaotic systems where transient instabilities—rather than asymptotic behavior—play a central role in the formation of extreme events~\cite{mohamad2016probabilistic}.

In the studies of nonlinear dynamical systems, Lyapunov spectral analysis has been used to characterize chaotic behaviors\cite{goldhirsch1987stability, ott2002chaos}. The asymptotic stability of the reference trajectory $\mathbf{u}(t)$ with respect to infinitesimal perturbations is classically characterized by the Lyapunov spectrum. Lyapunov exponents give a measure of the mean divergence rates of nearby trajectories on a strange attractor of the dynamical system, quantifying the long-time exponential growth or decay rates of perturbations in multiple directions. Positive values indicating instability and negative values indicating decay. In practice, computing the Lyapunov spectrum requires long-term integration of the variational equation~\eqref{eq:setup_variational} which is numerically challenging~\cite{brown1991computing, sandri1996numerical}. Most importantly, because of the long time character of Lyapunov spectrum, transient instabilities are not captured and therefore Lyapunov spectrum cannot be utilized for any time of prediction. To overcome this limitation, the variational equation can be used in the context of optimal time dependents, which is discussed in the next section.

\paragraph{Extreme events observable.}
Let $z(t) \in \mathbb{R}$ denote an observable — a scalar quantity of interest used to characterize extreme events along the system evolution. 
We characterize extreme events as rare, intermittent excursions of the observable to unusually large values. In practice, we fix an extreme event level $z^\star$ (e.g., chosen based on a high quantile of the observed time series or a physically motivated reference value) and label the system as being in an extreme state whenever $z(t)\ge z^\star$.
We assume that $z(t)$ is obtained from the system state; namely, there exists a mapping $\mathcal{G}:\mathcal{U}\to\mathbb{R}$ such that
\begin{equation}\label{eq:setup_qoi_map}
    z(t) = \mathcal{Z}\!\big(\mathbf{u}(t)\big).
\end{equation}

\paragraph{Precursor for upcoming extremes.}
Our objective is to identify a \emph{precursor} — a low-dimensional, time-dependent indicator derived from the system evolution — that provide early warning of extreme excursions of the observable $z(t)$. Specifically, we seek a precursor signal in the form
\begin{equation}\label{eq:setup_precursor}
    \pi(t)=\Pi\!\big(\mathbf{u}(t)\big),
    \qquad \text{or more generally} \qquad
    \pi(t)=\Pi\!\big(\mathbf{u}_{[t-\Delta,t]}\big),
\end{equation}
computed from information available up to time $t$ (instantaneous state or a short history window of duration
$\Delta\ge 0$), such that large values of $\pi(t)$ indicate that the observable will exceed the extreme-event threshold $z^\star$ at a prescribed lead time $\tau > 0$, i.e., at time $t+\tau$.

\paragraph{Available data.}
We assume access to time series data from one (or more) trajectories, sampled at times $\{t_k\}_{k=0}^{N}$. In
particular, we consider state snapshots $\{\mathbf{u}(t_k)\}$ from which the precursor signal $\pi(t_k)$ and quantity of interest $z(t_k)$ can be evaluated.

\subsection{Optimal Time-Dependent Modes}
\label{sec:OTD}

The framework of \emph{Optimal Time-Dependent} (OTD) modes~\cite{babaee2016minimization}, has been developed to construct a time-dependent orthonormal basis that adapts with the evolving dynamics while remaining sensitive to the finite-time dynamic instabilities. The OTD formulation uses the variational equation~\eqref{eq:setup_variational} and provides a time dependent, orthonormal basis, while still spanning the same flow-invariant subspaces as the solutions of the variational equation. This property ensures that transient instabilities can be captured in a numerically stable manner. Specifically, the first $r$ OTD modes $\{\mathbf{v}_i\}_{i=1}^r$ are defined through the constrained minimization problem~\cite{babaee2016minimization}
\begin{equation}\label{eq:otd-min}
    \arg \min_{\mathbf{\dot{v}_i}} \sum_{i=1}^r||\mathbf{\dot{v}}_i - \mathbf{L}\mathbf{v}||^2 \quad \text{subject to} \quad\langle \mathbf{v}_i, \mathbf{v}_j \rangle = \mathbf{I}_{r \times r},
\end{equation}
where $\langle \cdot, \cdot \rangle$ is a suitable inner product, $\|\cdot\|$ the induced norm, and $\mathbf{I}_{r \times r}$ the identity matrix of size $r$ ($1 \leq r \leq n$). The optimization in eq.~\eqref{eq:otd-min} is performed with respect to $\dot{\mathbf{v}}_i$ and not $\mathbf{v}_i$, therefore, the OTD modes are by construction the best approximation of the linearized dynamics in the subspace that they span. For the generic dynamical system of eq.~\eqref{eq:setup_dynsystem} and an $r$-dimensional OTD subspace, the evolution equation of the $i$th mode is:
\begin{equation}\label{eq:otd-modes}
\dot{\mathbf{v}}_i = \mathbf{L} \mathbf{v}_i - \sum_{k=1}^{r} \left( \langle \mathbf{L} \mathbf{v}_i, \mathbf{v}_k \rangle \mathbf{v}_k - \mathbf{\Phi}_{ik} \mathbf{v}_k \right),
\end{equation}
where $\boldsymbol{\Phi}$ is an arbitrary skew-symmetric matrix. Following~\cite{blanchard2019analytical}, the $\boldsymbol{\Phi}$ is selected as:
\begin{equation}\label{eq:phi}
\mathbf{\Phi}_{ik} =
\begin{cases}
-\langle \mathbf{L} \mathbf{v}_k, \mathbf{v}_i \rangle, & \text{for } k < i, \\
0, & \text{for } k = i, \\
\langle \mathbf{L} \mathbf{v}_i, \mathbf{v}_k \rangle, & \text{for } k > i.
\end{cases}
\end{equation}

With this choice of $\boldsymbol{\Phi}$, the evolution equation for the $i$th OTD mode takes the form:
\begin{equation}\label{eq:otd-final}
\dot{\mathbf{v}}_i = \mathbf{L} \mathbf{v}_i - \langle \mathbf{L} \mathbf{v}_i, \mathbf{v}_i \rangle \mathbf{v}_i - \sum_{k=1}^{i-1} \left( \langle \mathbf{L} \mathbf{v}_i, \mathbf{v}_k \rangle + \langle \mathbf{L} \mathbf{v}_k, \mathbf{v}_i \rangle \right) \mathbf{v}_k
\end{equation}

In this formulation we obtain a lower-triangular structure that can be solved sequentially by forward substitution. A key property of the OTD modes is their exponentially fast alingment with the transiently most unstable directions \cite{babaee2017reduced}. Specifically, the first OTD mode, $\mathbf{v}_1$, aligns with the most unstable direction, whereas the second mode, $\mathbf{v}_2$, is constrained to remain orthogonal to $\mathbf{v}_1$, spanning the second most unstable direction; together, they span the two-dimensional subspace exhibiting the fastest growth. The orthonormality of the OTD modes ensures numerical stability and provides a rigorous framework for analyzing finite-time instabilities, by continuously tracking the most unstable directions in phase space, even for time dependent systems. 

Under mild assumptions, the OTD subspace converges exponentially to the dominant eigenspace of the Cauchy–Green tensor, which characterizes transient instabilities \cite{babaee2017reduced}. At hyperbolic fixed points, OTD modes converge to the subspace spanned by the $r$ least-stable eigenvectors of the linearized operator $\mathbf{L}$ \cite{babaee2016minimization}. We also note that OTD modes coincide to Gram–Schmidt vectors, or backward Lyapunov vectors, which are classical tools for identifying unstable directions in phase space \cite{blanchard2019analytical}. 

\begin{figure}
    \centering
    \includegraphics[width=0.6\linewidth]{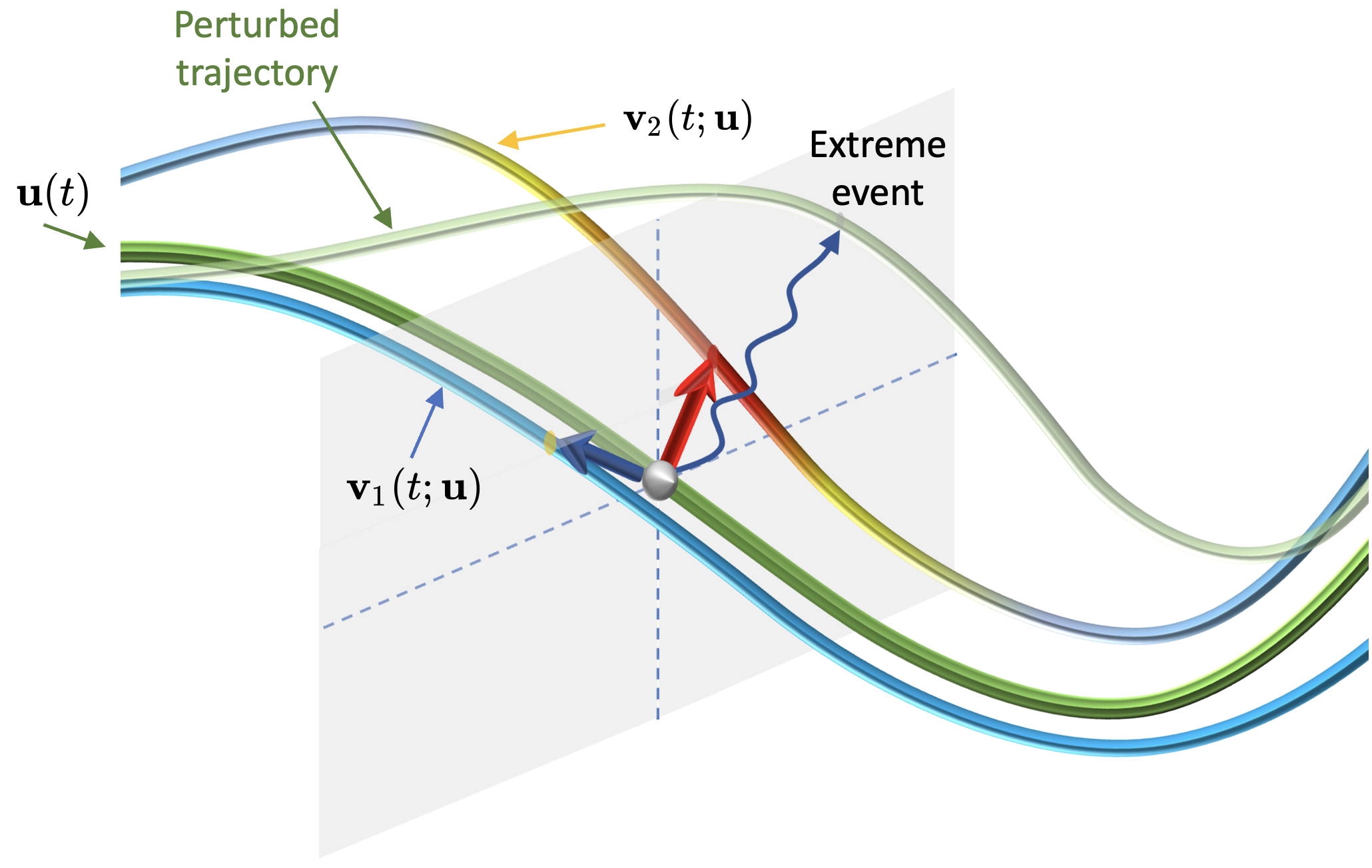}
    \caption{An illustration of the first two OTD modes, colored according to their stability properties (blue is stable and red is unstable), along a reference trajectory (green), $\mathbf{u}(t;\mathbf{u}_0)$.
    A perturbation generates a nearby trajectory (shown in light green color), which undergoes rapid growth along the first OTD direction, resulting in an extreme event.}
    \label{fig:otd_traj}
\end{figure}

Figure~\ref{fig:otd_traj} illustrates graphically the geometry of OTD modes for $r = 2$. The main trajectory around which the OTD modes are computed is shown with green color. The two OTD modes are colored according to their stability: blue indicates a stable direction and red an unstable direction. A perturbed trajectory (light green) is also shown as it undergoes a rapid growth towards the unstable (first OTD) direction, resulting in an extreme event. Using the stability properties along the direction of the OTD modes provides information which could serve the discovery of effective precursors for upcoming bursts in chaotic trajectories \cite{farazmand2016dynamical}.

\paragraph{Order reduction of the dynamics on OTD modes.} The OTD modes span flow-invariant subspaces of the tangent space,  enabling a dynamically consistent reduction of the linear operator $\mathbf{L}$ onto the OTD subspaces. In the case of infinite-dimensional setting, projection of the linearized operator to an $r$-dimensional OTD subspace yields a finite-dimensional reduced operator, i.e., an $r \times r$ matrix~\cite{farazmand2016dynamical}. We present the derivation in finite-dimensional form for clarity. Let $\boldsymbol{\xi} \in \mathbb{R}^n$ denote a solution of the full variational equation~\eqref{eq:setup_variational}, and let $\boldsymbol{\eta} \in \mathbb{R}^r$ denote its projection onto the OTD basis $\mathbf{V}$, 
\begin{equation}
    \boldsymbol{\eta} = \mathbf{V}^{\mathsf{T}}\boldsymbol{\xi},
\end{equation}
where 
$\mathbf{V}(t) =
\begin{bmatrix}
    \mathbf{v}_1 & \mathbf{v}_2 & \cdots & \mathbf{v}_r
\end{bmatrix}
\in \mathbb{R}^{n \times r}$ is the time-dependent matrix whose columns are the OTD modes obtained from eq.~\eqref{eq:otd-final}. The perturbation can equivalently be expressed as $\boldsymbol{\xi} = \mathbf{V}\boldsymbol{\eta}$.
Substituting this representation into the variational equation~\eqref{eq:setup_variational} yields the reduced-order linear equation
\begin{equation}\label{eq:red-eq}
    \dot{\boldsymbol{\eta}} = \Big( \mathbf{V}^{\mathsf{T}}\mathbf{L}\mathbf{V} \Big)\boldsymbol{\eta},
\end{equation}
which is dynamically consistent with the full system. Conversely, if $\boldsymbol{\eta}$ solves~\eqref{eq:red-eq}, then $\boldsymbol{\xi} = \mathbf{V}\boldsymbol{\eta}$ is an exact solution of the full variational equation (see \cite{babaee2016minimization}, Theorem~2.4). We therefore define the reduced linear operator $\mathbf{L}_r : \mathbb{R}^r \to \mathbb{R}^r$ as
\begin{equation}\label{eq:red-oper}
    \mathbf{L}_r = \mathbf{V}^{\mathsf{T}} \mathbf{L}\mathbf{V}. 
\end{equation}
A key advantage of this reduction is that it preserves the transient instabilities of the full-order system, regardless of whether they arise from modal or non-modal growth, assuming that $r$ is sufficiently large to capture the unstable subspace. Since the OTD basis evolves along the trajectory, it adapts to the most unstable directions encountered in phase space, making it a natural projection framework. However, as discussed in~\cite{farazmand2016dynamical}, the eigenvalues of $\mathbf{L}_r$ cannot be interpreted as physical growth or decay rates. Instead, one may consider the symmetric part
\begin{equation}\label{eq:sym-oper}
    \mathbf{S}_r := \tfrac{1}{2}\big(\mathbf{L}_r + \mathbf{L}_r^{\mathsf{T}}\big),
\end{equation}
whose eigenvalues $\lambda_1 \geq \lambda_2 \geq \cdots \geq \lambda_r$ provide instantaneous measures of perturbation growth or decay within the OTD subspace. Even this measure of growth is not effective though and for extreme events precursors one should focus on finite-time analysis, i.e. the use of finite-time Lyapunov exponents.

\subsection{Finite-Time Lyapunov Exponents}
\label{sec:ftle}


The finite-time Lyapunov exponents (FTLEs) quantify the growth or decay of infinitesimal perturbations along the system's trajectory over a finite time window, providing a measure of transient instability of the dynamics~\cite{abarbanel1991variation}. Unlike asymptotic Lyapunov exponents, which reflect long-time averaged behavior, FTLEs capture time-dependent and state-dependent amplification of disturbances. 
In the present setting, we assume that the system is observed at the current time $t$, and we are interested in characterizing the cumulative amplification of perturbations over the preceding interval $[t-T, \,t]$, where $T > 0$. An infinitesimal perturbation $\boldsymbol{\xi}_{t-T}$ applied at time $t-T$ evolves forward to the current time $t$ according to the linearized flow operator 
\begin{equation}
    \Psi_{t-T}^{\,t}:= D_{\mathbf{u}}\psi_{t-T}^{\,t}(\mathbf{u}), 
\end{equation}
where $\psi_{t-T}^{\,t}$ denotes the flow map defined in Eq.~\eqref{eq:flow_map}. Consequently, the perturbation at time $t$
satisfies
\begin{equation}
    \boldsymbol{\xi}(t)
    =
    \Psi_{t-T}^{\,t}\,(\boldsymbol{\xi}_{t-T}).
\end{equation}

To measure the growth of the infinitesimal perturbations in phase space~\cite{haller2002lagrangian}, the \textit{right Cauchy--Green deformation tensor} is typically used 
\begin{equation}\label{eq:Cauchy-Green}
    \mathbf{C}_{t-T}^{\,t} 
    = \left[\Psi_{t-T}^{\,t}\right]^{\!\top}
      \Psi_{t-T}^{\,t},
\end{equation}
which is symmetric and positive definite, and its eigenvalues $\lambda_i(t; \,t-T)$ quantify the finite-time stretching of infinitesimal perturbations along orthogonal directions in phase space. We order the Cauchy-Green eigenvalues in a descending order, 
\begin{equation}
    \lambda_1 \geq \lambda_1 \geq \ldots \geq \lambda_n \geq0.
\end{equation}
The \textit{finite-time Lyapunov exponents} over the interval $[t-T,\,t]$ are defined as
\begin{equation}\label{eq:FTLE}
    \Lambda_i(t; \, t-T)
    = \frac{1}{T}\,
      \log\!\sqrt{\lambda_i(t; \,t-T)},
    \qquad i = 1,\dots,n,
\end{equation}
Large values of $\Lambda_i$ indicate strong local stretching and sensitivity to initial conditions, whereas during quiescent phases their magnitudes remain small, reflecting weak perturbation growth. As the system approaches an extreme or dissipative event, the FTLEs exhibit sharp peaks that signal the transient amplification of instabilities preceding the onset of such bursts.

\subsubsection{Reduced-order computation of FTLE}
\label{sec:reduced-ftle}

The direct computation of FTLEs in high-dimensional systems is computationally prohibitive, as it requires evaluating the full Cauchy--Green tensor - in practice this means the solution of at least $n$ perturbed trajectories that will be used to quantify the the Cauchy-Green tensor. To address this limitation, Babaee et al.~\cite{babaee2017reduced} developed a reduced-order framework for computing FTLEs utilizing the subspace spanned by OTD modes, which adapt dynamically to transient instabilities. It was proved that, under suitable conditions, the OTD modes converge exponentially fast to the dominant eigendirections of the Cauchy--Green tensor corresponding to the strongest finite-time instabilities, i.e., those associated with the largest FTLEs. We summarize the reduced-order procedure for computing FTLEs over the finite-time interval \([t-T,\,t]\).

\begin{enumerate}
    \item \textbf{Trajectory.}  
    Utilize the given data, expressed through a trajectory $\mathbf{u}(t)$ over the interval $ [t-T,\, t]$.

    \item \textbf{OTD subspace construction.}  
    Compute the $r$-dimensional OTD basis corresponding to this trajectory using Eq.~\eqref{eq:otd-final}.

    \item \textbf{Evolution of the reduced-order fundamental matrix.}  
    Evolve the reduced fundamental solution matrix $\mathbf{Y}_{t-T}^{t} \in \mathbb{R}^{r \times r}$ according to
    \begin{equation}\label{eq:Y_evol}
        \frac{d}{dt}\,\mathbf{Y}_{t-T}^{t} = \mathbf{L}_r\,\mathbf{Y}_{t-T}^{t},
        \qquad \mathbf{Y}_{t-T}^{t} = I_r, 
    \end{equation}
    where $\mathbf{L}_r$ denotes the projection of the full linearized operator $\mathbf{L}$ onto the OTD subspace; see Eq.~\eqref{eq:red-oper}. Here, $I_r$ denotes the identity matrix in $\mathbb{R}^{r \times r}$.
    
    \item \textbf{Reduced-order Cauchy--Green tensor.}  
    Construct the reduced-order right Cauchy--Green tensor
    \begin{equation}
        \mathbf{R}_{t-T}^t = \left(\mathbf{Y}_{t-T}^{\,t}\right)^{\!\top}\,\mathbf{Y}_{t-T}^{\,t},
    \end{equation}
    with eigenvalues $\gamma_1 \geq \gamma_2 \geq \cdots \geq \gamma_r\geq0$.  

    \item \textbf{Reduced-order FTLEs.}  
    The finite-time Lyapunov exponents in the reduced subspace over the interval $[t-T,\,t]$ are given by
    \begin{equation}\label{eq:red-FTLE}
       \Gamma_i(t; \,t-T) 
       = \frac{1}{T}\,\log \sqrt{\gamma_i(t; \,t-T)},
       \qquad i = 1,\ldots,r.
    \end{equation}
\end{enumerate}



\section{Extreme Event Precursors}
\label{sec:problem}
\begin{figure}
    \centering
    \includegraphics[width=\linewidth]{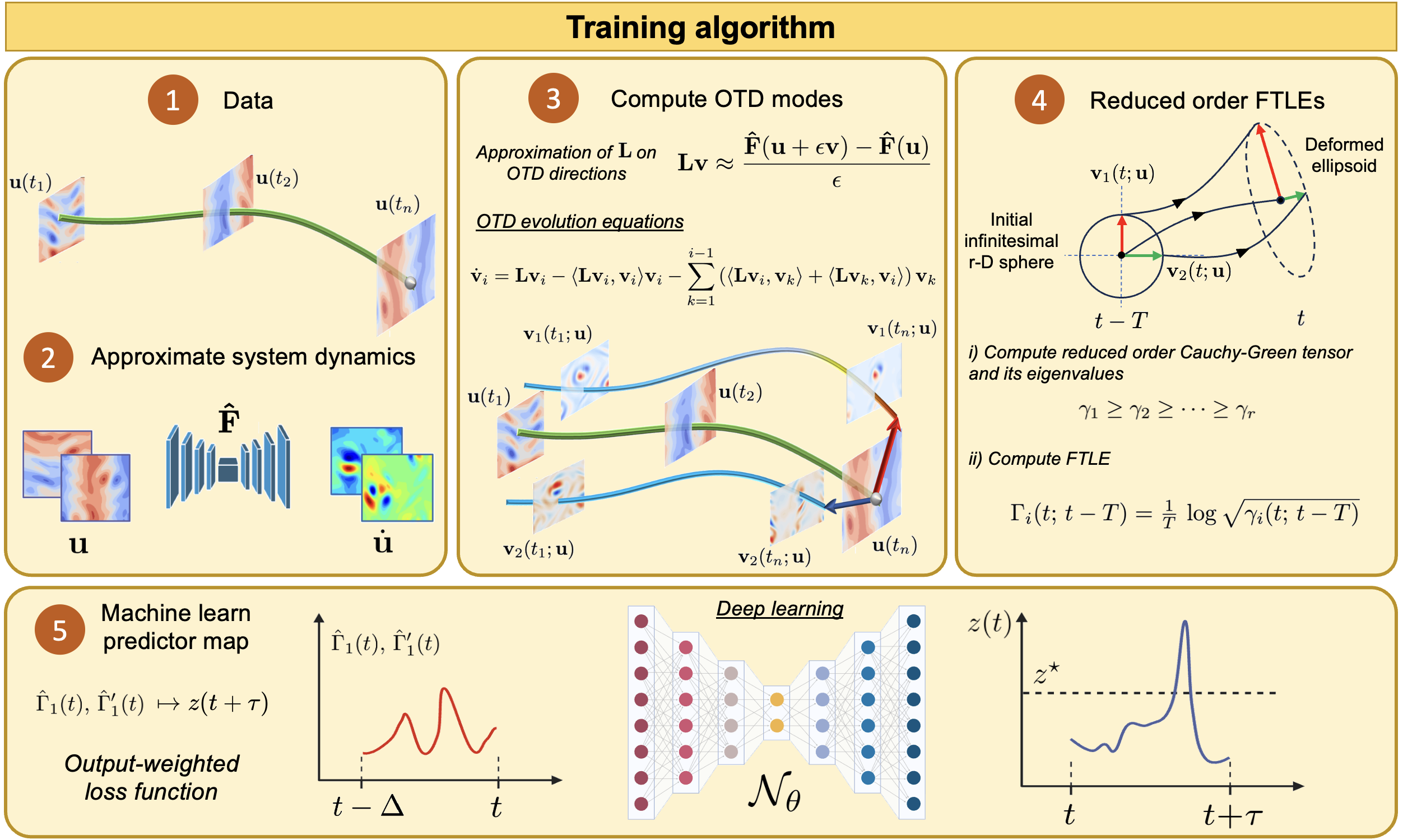}
    \caption{Illustration of the training algorithm. A long sequence of snapshots of the system state allows for the approximation of the dynamics (step 2). Dynamics is used to approximate the action of the linearized flow on the OTD subspace, which allows for parsimonious evolution of the OTD modes (step 3). A computation of the FTLE is performed, within the OTD subspace (step 4). Final step 5 is the machine learning of a map from the dominant FTLE to the predicted observable for extreme events.}
    \label{fig:Methodology}
\end{figure}

We now formulate a fully data-driven framework for the real-time prediction of extreme events in chaotic, high-dimensional dynamical systems. The framework consists of two components. The first component provides a methodology for computing reduced-order FTLEs using just data, i.e the sequence of snapshots $\{\mathbf{u}(t_k)\}_{k=0}^N$. The second component is the predictive, where the leading FTLE, ${\hat{\Gamma}_1}$, computed from system observations up to time $t$, provides dynamical information that is mapped directly to the predicted value of the observable, $\hat{z}$, at a prescribed lead time $t+\tau$. 
If the predicted value exceeds the threshold condition $\hat{z}(t+\tau) \geq z^\ast$, we have have identified an extreme event. In this section we discuss in detail the various steps involved (Figure \ref{fig:Methodology}).


\subsection{Dynamics and OTD modes from data}
We assume an equation-agnostic setup where only discrete time series of the system state is available. To compute OTD modes, it is essential to obtain an approximation of the variatonal equation. This is achieved by first modeling in a data-driven way the system dynamics, $\mathbf{\hat F}$ (Figure \ref{fig:Methodology} - step 2). Here the hat denotes the approximation of the dynamical system. 

A broad class of methods exists for inferring the dynamical system, $\mathbf{\hat F}$, directly from data~\cite{vlachas2018data, wan2018data, kovachki2023neural, yu2024learning}. Here we assume that the snapshots are sampled along long trajectories with a uniform and sufficiently small sampling time-step $\Delta t$, so we approximate $\mathbf{\hat F}(\mathbf{u})$ utilizing a forth-order central finite-time-differences scheme. See Appendix A.1 for details of this step.

To evolve OTD modes it is sufficient to compute the action of the linearized operator $\mathbf{L}(\mathbf{u})$ just on the OTD modes. We employ the widely used practice of estimating Jacobian-vector products through finite differences of the vector field~\cite{an2011finite}. This matrix-free approach avoid reconstructing or storing the Jacobian (linearized operator), which is critical for high-dimensional systems. Specifically, the directional derivative of $F(\mathbf{u})$ along $\mathbf{v}$ is approximated as
\begin{equation}\label{eq:fin-diff}
    \mathbf{Lv} \approx \frac{\mathbf{\hat F}(\mathbf{u} + \epsilon\mathbf{v}) - \mathbf{\hat F}(\mathbf{u})}{\epsilon},
\end{equation}
where $\epsilon \in \mathbb{R} $ is a small finite-difference step, which allows us to compute the action of the linearized operator $\mathbf{L}$ on the OTD modes $\mathbf{v}$. With these ingredients in place, we compute the OTD eq.~\eqref{eq:otd-final} (Figure \ref{fig:Methodology} - step 3) and the reduced-order linearized operator $\widehat{\mathbf{L}}_r$ using the projection formula~\eqref{eq:red-oper} in a fully data-driven way. 

It is important to emphasize that the learned dynamical system, $\mathbf{\hat F}(\mathbf{u})$, is never directly used for prediction or forecast of extreme events. It only employed to approximate the variational flow, and from there the OTD modes. In other words the approximated dynamics are only used to characterize the local neighborhood of the system state and its transient instabilities.

\subsection{Mapping FTLEs to extremes}
We are now in position to compute FTLEs, using the reduced-order linearized operator, $\widehat{\mathbf{L}}_r$, and the process described in Section \ref{sec:reduced-ftle} (Figure \ref{fig:Methodology} - step 4). We focus on the leading FTLE, $\hat{\Gamma}_1(t)$, and machine learn a model that performs long-horizon, $\tau$, prediction of extreme events, quantified through the observable, $z(t+\tau)$ (Figure \ref{fig:Methodology} - step 5). 

The predictive capability of the model in capturing such extremes is evaluated through binary classification metrics that differentiate between extreme and non-extreme events, supplemented by conditional statistical measures that assess the forecasting skill of the precursor in the vicinity of extreme occurrences.

\subsubsection{Deep learning a precursor model}
\label{subsec:time-ser-for}
The objective of this step is to forecast the future evolution of the observable \(z(t)\) for times $t \in [t,\,t+\tau]$, given the value of the dominant FTLE, $\hat{\Gamma}_1(t)$. In addition to the value of the FTLE we will also include its time derivative
$\hat{\Gamma}'_1(t)$. The inclusion of the derivative provides information on the instantaneous growth rate of local instabilities, enriching the temporal context available to the forecasting model form a two-channel input sequence for the forecasting model:
\[
\mathbf{\pi}(t) = [\,\hat{\Gamma}_1(t),\, \hat{\Gamma}'_1(t)\,].
\]

The problem is formulated as a sequence-to-sequence learning task, where a nonlinear operator is trained to map the recent history of the precursor to the corresponding future dissipation response over a prediction horizon~$[t, \, t + \tau]$. The lookback window of length~\(\Delta\), partinioded over $n_\Delta$ steps, provides the model with sufficient temporal context to capture both the amplitude and rate of change of the system’s instability, while length~\(\tau\) and its partition to $n_\tau$, specifies the number of future steps to be predicted. To implement this forecast step we employ a Transform architecture (details in Appendix A.2).

\begin{equation}
\mathcal{N}_\theta:\; \mathbb{R}^{2 n_\Delta} \longrightarrow \mathbb{R}^{n_\tau}, 
\qquad 
\hat{\mathbf{Z}} = 
\mathcal{N}_\theta\!\left(\boldsymbol{\
\Pi}\right),
\label{eq:seq2seq}
\end{equation}
where
\[
\boldsymbol{\Pi}
= [\,\pi(t-\Delta), \ldots, \pi(t)\,],
\quad
\hat{\mathbf{Z}}
= [\,\hat{z}(t+\tau/n_\tau), \ldots, \hat{z}(t+\tau)\,].
\]
The \(\mathcal{N}_\theta\) denotes the machine-learning model parameterized by~\(\theta\), trained to approximate the nonlinear mapping between the  temporal evolution of the precursors $\pi(t)$ and the observable $z(t)$.

\paragraph{Loss Function.}
To train the prediction model, we employ an \emph{output-weighted loss function} designed to emphasize the accurate prediction of rare, high-magnitude events. 
Following the concept of output-weighted regression~\cite{rudy2023output, barthel2023harnessing}, we employ a weighting scheme inversely proportional to the probability density of the target variable, thereby amplifying the influence of rare events on the total loss. 
The \emph{output-weighted mean absolute error} (MAE\(_{\text{OW}}\)) loss is defined as
\begin{equation}
    \mathcal{L}_{\text{OW}} 
    = \mathbb{E}_{z}\!\left[
        \frac{|\,\hat{z} - z\,|}{p_{z}(z)}
      \right],
    \label{eq:owmae}
\end{equation}
where \(p_{z}(z)\) denotes the probability density function (PDF) of the true output and acts as a weighting function that adjusts the contribution of each sample according to its rarity. 
In practice, the empirical form of Eq.~\eqref{eq:owmae} is estimated as
\[
\mathcal{L}_{\text{OW}}
    = \frac{1}{N}\sum_{j=1}^{N}
      \frac{|\,\hat{z}_j - z_j\,|}{p_{z}(z_j)},
\]
where \(z_j\) and \(\hat{z}_j\) denote the true and predicted values of the observable at sample \(j\), and \(p_{z}(z_j)\) is estimated via kernel density estimation (KDE) from the training data. 

This formulation increases the penalty for errors in regions where the probability density \(p_{z}(z_j)\) is small, directing the optimization toward better prediction of rare and extreme events. 
Since each sample’s contribution to the loss is scaled by \(1/p_{z}(z_j)\), values with smaller probability (i.e., smaller denominators) lead to larger loss terms, while frequent, nominal states contribute less. 

\subsection{Summary of the prediction algorithm}
\begin{figure}
    \centering
    \includegraphics[width=\linewidth]{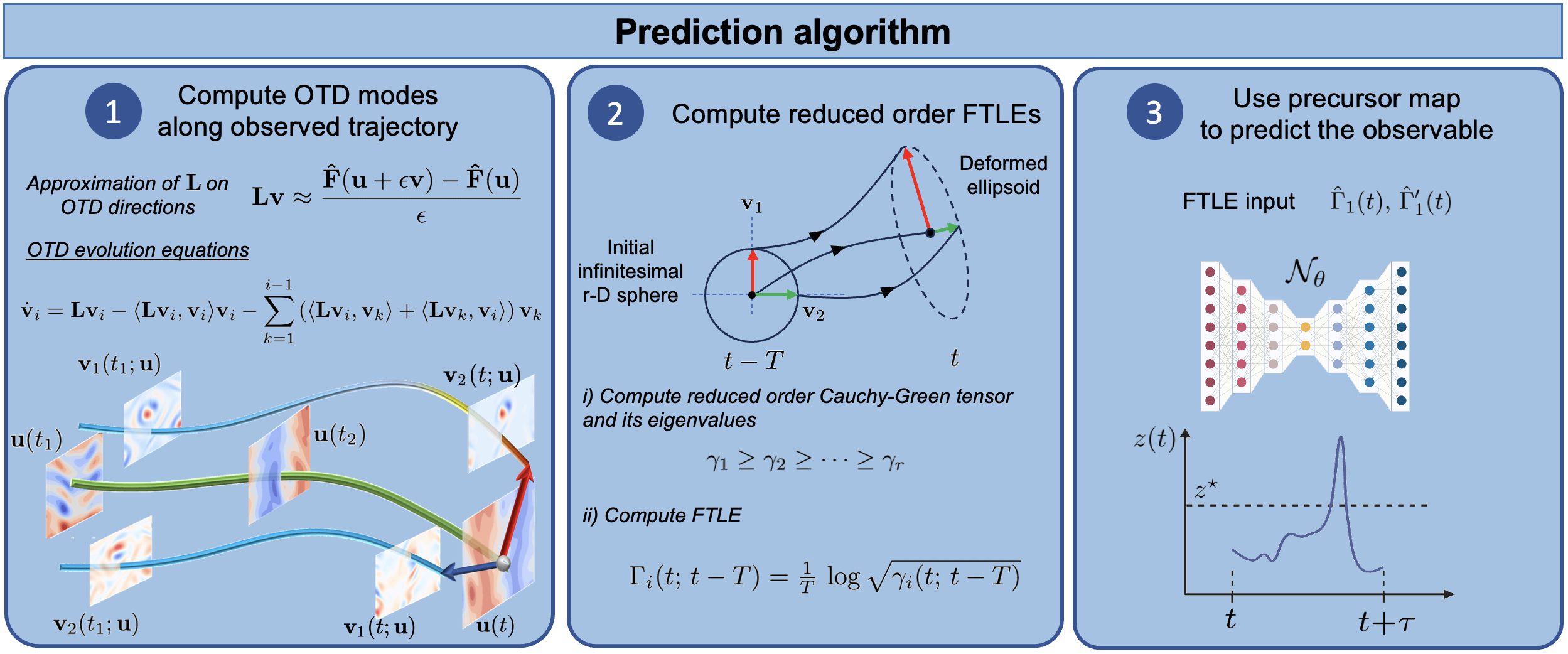}
    \caption{Illustration of the prediction  steps: 1. Computation of OTD modes; 2. Computation of the associated FTLEs; 3. Prediction of the observable of interest for extreme events.}
    \label{fig:Prediction}
\end{figure}

The prediction algorithm can thus be summarized with the following steps (Figure \ref{fig:Prediction}):
\begin{enumerate}
    \item Use the history of the system state, $\mathbf{u}(t)$, and compute the OTD modes, $\mathbf{v}_1(t;\mathbf{u}), ..., \mathbf{v}_r(t;\mathbf{u})$ up to the current time, $t$. 
    \item Compute the FTLEs, $\Gamma_i(t;\,t-T), i=1,...,r$, up to to the current time, $t$, starting from $t-T$ and over a finite-time horizon of length $T$.
    \item Use the prediction map to compute the value of the observable over the prediction horizon, $z(t+\tau)$.
\end{enumerate}

\subsection{Performance measures for precursors of extreme events}
\label{sec:measures}
To evaluate the predictive skill of the proposed precursors, we adopt several binary classification metrics~\cite{guth2019machine}. Specifically, each prediction outcome is categorized as one of four outcomes: a \textit{true positive} (TP) when an extreme event is both observed and correctly predicted, i.e., $z > z^\star$ when $\hat{z} > \hat {z}^\star$; a \textit{true negative} (TN), when a non-extreme event is correctly identified, i.e., $z < z^\star$ and $\hat{z} < \hat{z}^\star$; a \textit{false positive} (FP), when an extreme is predicted but does not occur, i.e., $z < z^\star$ and $\hat{z} >\hat {z}^\star$; and a \textit{false negative} (FN), when an actual extreme event is missed by the model, i.e., $z > z^\star$ and $\hat{z} < \hat z^\star$. To assess the accuracy of the precursors, we employ the following criteria which are effective to deal with the strongly unbalanced character of the datasets that contain extreme rare events:\\

1) \textit{F1-score} provides a unified quantitative metric that balances the precursor's ability to both correctly identify and accurately predict extreme events. It combines two complementary measures: the \textit{precision}, which denotes the probability that an event predicted as extreme is indeed a true extreme (reflecting the model’s reliability in avoiding false alarms), and the \textit{recall}, which denotes the probability that an event that is truly extreme is correctly identified as such (reflecting the model’s ability to capture all extreme occurrences). These quantities are defined as
\[
S (\hat z^\star) = \frac{\text{TP}(\hat z^\star)}{\text{TP} (\hat z^\star)+ \text{FP}(\hat z^\star)} \quad \text{[Precision]}, \quad R(\hat z^\star) = \frac{\text{TP}(\hat z^\star)}{\text{TP}(\hat z^\star) + \text{FN}(\hat z^\star)} \quad \text{[Recall]}
\]
The F1-score is computed as the harmonic mean of precision and recall:
\[
\text{F1} = 2 \times \frac{S \times R}{S + R}
\]
This formulation penalizes models that achieve high performance on only one of the two metrics and attains its maximum value, $\text{F1} = 1$, when both precision and recall are perfect (i.e., the model neither generates false extremes nor misses true ones). However, the F1 score depends explicitly on the chosen threshold $\hat z^\star$.\\

2) \textit{Area under the Precision-Recall Curve (AUC)}: 
To mitigate the dependence of the evaluation metrics on the prediction threshold $\hat z^\star$, we adopt a threshold-independent measure by integrating precision and recall over the entire range of $\hat z^\star$. Specifically, we first fix the extreme-event threshold $z^\star$ used to label true events and construct the \textit{precision–recall (PR) curve} by varying the prediction threshold $\hat z^\star$ and plotting precision as a function of recall. This curve provides a comprehensive view of how the model’s classification performance varies as the decision boundary changes. The \textit{area under the PR curve} (AUC) is then defined as
\begin{equation}
\label{eq:AUC}
\text{AUC} = \int_0^1 S(R)\, dR = \int_{-\infty}^{\infty} S(\hat z^\star) \left|\frac{\partial R}{\partial \hat z^\star}\right| d\hat z^\star,
\end{equation}
where $S$ denotes precision as a function of recall $R$.  

If $\hat z^\star$ is set too low, most extreme events will be correctly identified (high recall) but many false positives will occur (low precision). Conversely, if $\hat z^\star$ is too high, false positives are minimized but numerous true extremes are missed (low recall). A robust predictor achieves simultaneously high precision and recall over a wide range of $\hat z^\star$ values. The AUC thus provides a scalar, threshold-independent measure of this robustness, with larger AUC values (closer to 1) indicating more consistent and reliable detection performance.\\

3) \textit{Maximum adjusted area under the curve}, $\alpha^*$: Following \cite{guth2019machine}, the \textit{maximum adjusted area under the curve} criterion compares the AUC associated with an extreme-event rate $\omega$, denoted $\text{AUC}(\omega)$, to that of an uninformed (random) predictor whose expected AUC equals $\omega$. The difference $\text{AUC}(\omega) - \omega$ quantifies the gain in predictive skill relative to random guessing, and the maximum of this difference over all possible event rates defines
\begin{equation}\label{eq:alpha-star}
\alpha^* = \max_{\omega \in [0,1]} \big(\text{AUC}(\omega) - \omega\big).
\end{equation}
The quantity $\alpha^*$ identifies the event rate at which the predictor achieves the greatest improvement over a random classifier, highlighting the regime where the model most effectively separates extreme from quiescent states. Unlike the standard AUC, this measure is fully threshold-independent and particularly suitable for evaluating predictors in highly unbalanced datasets containing rare events.\\

4) \textit{Extreme Event Count}: To assess the model’s ability to reproduce the temporal occurrence of extreme events, we employ the \textit{extreme event count} approach. This measure quantifies how well the model captures the number of distinct extreme events within a given time window. Formally, a time instant $t_j$ is classified as an extreme event, denoted $t_{EE}$, if it satisfies
\[
t_{EE}: \quad t_j \ \text{s.t.} \ \left[\frac{\partial z}{\partial t}\bigg|_{t_j} = 0 \quad \text{and} \quad z(t_j) > z^\star \right].
\]
The total number of extreme events occurring within the interval $[t_1, t_2]$ is then given by
\[
N_{EE}(t_1, t_2) = \sum_{j=j_1}^{j_2} \delta_{t_j, t_{EE}},
\]
where $\delta_{t_j, t_{EE}} = 1$ if $t_j$ corresponds to an identified extreme event and $0$ otherwise. To prevent spurious detections due to high-frequency noise, a minimum temporal separation between successive peaks is imposed, defined as the characteristic period associated with the dominant extreme-event frequency, $T_{EE} = 1/f_{EE}$. While the approach depends on two user-defined parameters---the threshold $z^\star$ and the minimum separation $T_{EE}$---the extreme event count provides a direct and interpretable measure of the model’s forecasting skill. 

In order to compare the predicted and true counts of extreme events we evaluate the absolute difference,
\[
\Delta N_{EE} = |N_{EE}^{\text{true}} - N_{EE}^{\text{pred}}|,
\]
as an indicator of how accurately the model reproduces the true frequency of extreme occurrences over time. Smaller $\Delta N_{EE}$ values indicate better agreement between the predicted and observed event statistics.

\paragraph{Quantification of the tail statistics}
To assess how closely the learned model reproduces the true probability density function $p_z$, we compare their distributions, with particular attention to the behavior of the tails. Following~\cite{rudy2023output}, we employ a metric that measures the average absolute difference between the logarithms of the true and predicted densities over the intersection of their respective supports. The metric is normalized by the size of this intersection, thereby penalizing cases where the overlap between the two distributions is limited. It is defined as
\begin{equation}
\label{eq:metric-D}
    \mathbb{D}(p_z, \hat{p}_z) = \frac{1}{|\Omega(p_z, \hat{p}_z)|^2} \int_{\Omega(p_z, \hat{p}_z)} |\log(p_z(z)) - \log(\hat{p}_z(z))| dz,
\end{equation}
where,
\[
\Omega(p_z, \hat{p}_z) \approx \text{supp}(p_z) \cap  \text{supp}(\hat{p}_z) ,
\]
and $\hat{p}_z$ denotes the density estimated from the learned model. Because both $p_z$ and $\hat{p}_z$ are empirically approximated from finite data, their exact support is unknown, and the behavior of low-density regions is difficult to capture accurately. Nevertheless, since $\mathbb{D}$ is sensitive to both the magnitude and extent of overlap, we approximate each distribution’s support as the interval spanning the observed data range. While this underestimates the true width of the support, it provides a consistent and practical approach for computing $\mathbb{D}$.

\section{Kolmogorov Flow as a Prototype Model for Extreme Events}
\label{sec:Kolmogorov}

To demonstrate the developed scheme we employ the Kolmogorov flow. The two-dimensional Kolmogorov flow is a canonical solution of the incompressible Navier--Stokes equations subject to a sinusoidal body force  \cite{foias2001navier}. The governing equations are  
\begin{equation}\label{eq:nse}
    \frac{\partial \mathbf{u}}{\partial t} 
    = -\mathbf{u} \cdot \nabla \mathbf{u} - \nabla p 
      + \nu \nabla^2 \mathbf{u} + \mathbf{f}, 
    \qquad \nabla \cdot \mathbf{u} = 0,
\end{equation}
where $\mathbf{u}(\mathbf{x},t) \in \mathbb{R}^2$ is the velocity field, $p(\mathbf{x},t)$ is the pressure, and $\nu = 1/Re$ is the kinematic viscosity, inversely proportional to the Reynolds number. The external forcing is taken to be a sinusoidal shear in the $x$-direction,  
\[
    \mathbf{f}(\mathbf{x}) = \sin(n y)\,\mathbf{e}_1, 
    \qquad \mathbf{e}_1 = (1,0)^{\mathsf{T}}, \quad n \in \mathbb{N}.
\]

The domain is the two-dimensional torus $\mathbb{T}^2 = [0,2\pi]^2$ with periodic boundary conditions. The solution is given by the time-dependent velocity--pressure pair $(\mathbf{u},p)$.  

For forcing wavenumber $n = 1$, the Kolmogorov flow admits a stable laminar solution for all Reynolds numbers \cite{foias2001navier}. In contrast, for $n > 1$ and sufficiently large Reynolds numbers, the laminar solution loses stability. As demonstrated in \cite{platt1991investigation, chandler2013invariant}, the flow undergoes a transition to spatiotemporal chaos. The resulting turbulent attractor has a high-dimensional structure, with its dimension growing approximately linearly with the Reynolds number. This property renders the analysis of intermittency in turbulent flows particularly challenging. In the present study, we focus on the case $n = 4$ and $Re = 40$, where the Kolmogorov flow exhibits chaotic dynamics and evolves on a strange attractor.

Important properties of the Kolmogorov flow are the energy input $I$, the energy dissipation $D$, and the kinetic energy $E$. These quantities satisfy the energy balance law $dE / dt = I - D$, and are defined as:

\begin{equation}\label{eq:input-energy}
    I(t) = \frac{1}{L^2}\int_{\Omega} \mathbf{u}(\mathbf{x},t) \cdot \mathbf{f}(\mathbf{x}) \, d\mathbf{x},
\end{equation}
\begin{equation}\label{eq:dissipation}
    D(t) = \frac{\nu}{L^2}\int_{\Omega} |\omega(\mathbf{x},t)|^2 \, d\mathbf{x},
\end{equation}
\begin{equation}\label{eq:net-energy}
    E(t) = \frac{1}{2L^2}\int_{\Omega} |\mathbf{u}(\mathbf{x},t)|^2 \, d\mathbf{x},
\end{equation}
where $L = 2\pi$ is the side length of the domain $\Omega = [0,L]^2$ and $\omega = \nabla \times \mathbf{u}$ is the scalar vorticity field in two dimensions.

\begin{figure}[h]
    \centering
    \includegraphics[width=0.6\linewidth]{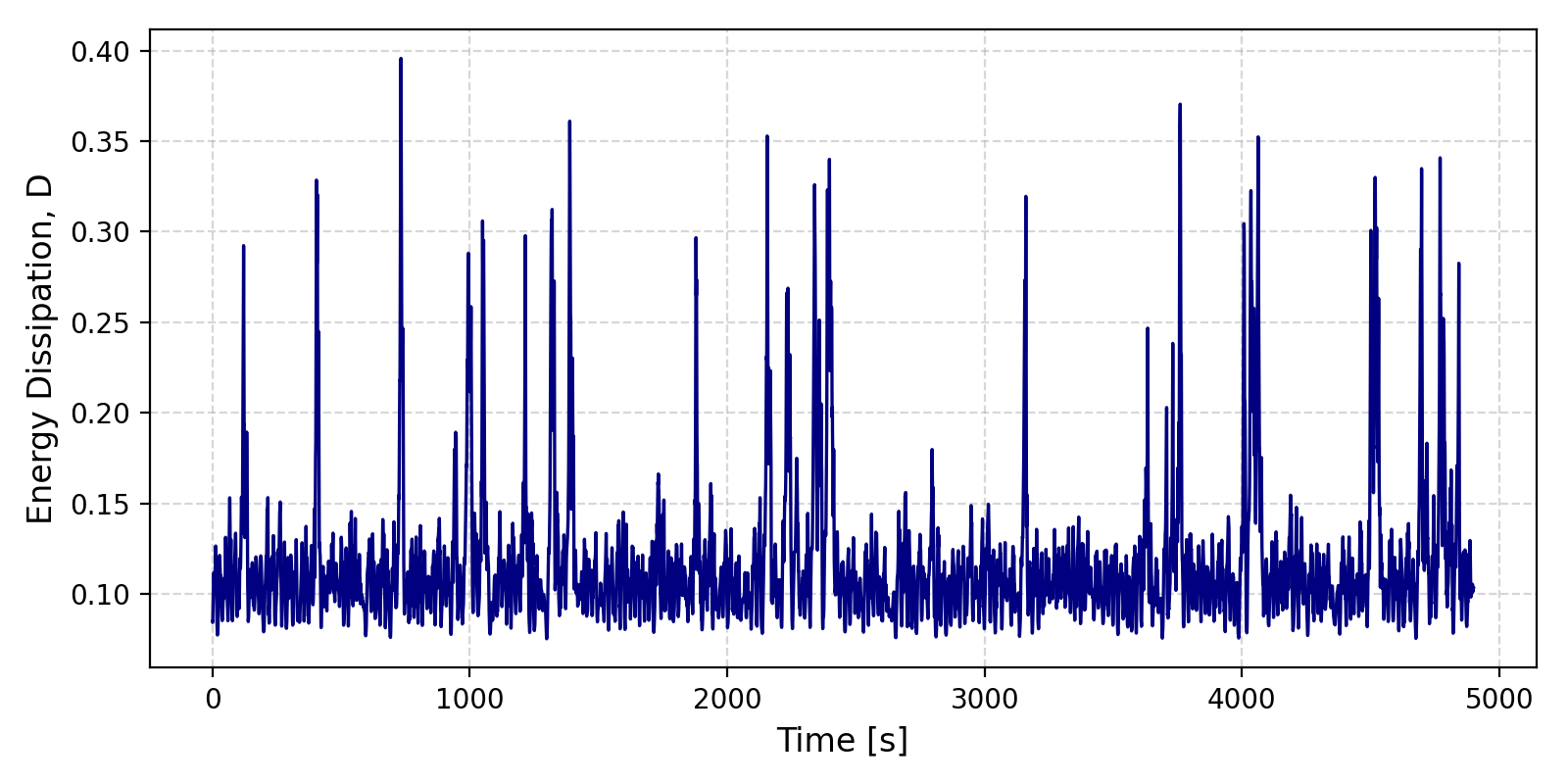}
    \caption{Time evolution of the energy dissipation $D(t)$ along a trajectory of Kolmogorov flow with $n = 4$ and $Re = 40$. The signal exhibits small-amplitude background oscillations around $D \approx 0.1$, punctuated by intermittent burst-like excursions corresponding to extreme dissipation events.}
    \label{fig:diss-energy}
\end{figure}

A ubiquitous feature of turbulent fluid flows is intermittency, manifested as sudden burst-like excursions in observable quantities. In this work, the energy dissipation $D(t)$ is the primary observable and the quantity of interest used to track extreme events, as illustrated in Figure~\ref{fig:diss-energy}. We define extreme events as instances in which $D(t)$ exceeds a threshold equal to two standard deviations above its mean. According to eq.~\eqref{eq:input-energy}, increases in the energy input $I(t)$ arise from transient alignment between the velocity field $\mathbf{u}$ and the external forcing $\mathbf{f}$. Such alignment produces sharp surges in $I(t)$, which, through the energy balance relation, translate into corresponding peaks in the energy dissipation $D(t)$. These observations indicate that the onset of extreme dissipation events is associated with the growth of perturbations that become transiently aligned with the forcing direction.

\subsection{OTD modes and the Kolmogorov flow}  
\label{sec:OTD-Kol}
The governing equations of the Kolmogorov flow can be written in projected form as:
\begin{equation}\label{eq:proj-NSE}
\frac{\partial \mathbf{u}}{\partial t} 
= \mathbf{F}(\mathbf{u})
= \mathbb{P}\left[-\mathbf{u} \cdot \nabla \mathbf{u}
+ \nu \nabla^2 \mathbf{u} + \mathbf{f}\right],
\end{equation}
where $\mathbb{P}$ denotes the Leray projection onto the divergence-free subspace, enforcing $\nabla\cdot\mathbf{u}=0$ and eliminating the pressure term so that the dynamics are entirely described by the velocity field. 
Within this setting, the nonlinear operator $\mathbf{F}:\mathcal{U}\to\mathcal{U}$ governs the time evolution of the velocity field according to the projected Navier--Stokes dynamics.

While in the developed prediction framework the OTD equations are fully-driven, just for the purpose of numerical comparison, we derive the variational equations for Navier-Stokes. Specifically, linearizing $\mathbf{F}$ about a state $\mathbf{u}$ yields the linearized Navier--Stokes operator $\mathscr{L}_{\mathrm{NS}}(\mathbf{u})$, which acts on a perturbation field $\mathbf{v}$ as,
\begin{equation}\label{eq:linearized_NS}
\mathscr{L}_{\mathrm{NS}} 
= \mathbb{P}\left[-\mathbf{u} \cdot \nabla \mathbf{v}
- \mathbf{v} \cdot \nabla \mathbf{u}
+ \nu \nabla^2 \mathbf{v}\right].
\end{equation}
The Leray projection $\mathbb{P}$ again ensures that the perturbation dynamics remain within the divergence-free subspace.
Substituting $\mathscr{L}_{\mathrm{NS}}(\mathbf{u}; \mathbf{v})$ into the OTD evolution equation~\eqref{eq:otd-final} yields the evolution of the $i$th OTD mode,
\begin{equation}\label{eq:otd-nse}
    \dot{\mathbf{v}}_i = \mathscr{L}_{\mathrm{NS}}(\mathbf{u}; \mathbf{v}_i) 
    - \langle \mathscr{L}_{\mathrm{NS}}(\mathbf{u}; \mathbf{v}_i), \mathbf{v}_i \rangle \mathbf{v}_i 
    - \sum_{k=1}^{i-1} \Big[ \langle \mathscr{L}_{\mathrm{NS}}(\mathbf{u}; \mathbf{v}_i), \mathbf{v}_k \rangle 
    + \langle \mathscr{L}_{\mathrm{NS}}(\mathbf{u}; \mathbf{v}_k), \mathbf{v}_i \rangle \Big] \mathbf{v}_k,
\end{equation}

We initialize the OTD modes as
\begin{equation}
    \mathbf{v}_i(\mathbf{x}, 0) = \frac{1}{\pi \sqrt{2}} 
    \begin{pmatrix}
        \sin(iy) \\
        0
    \end{pmatrix}, 
    \qquad i = 1, \dots, r,
\end{equation}
which are divergence-free, mutually orthogonal, and normalized in the $L^2$ sense. This initialization is performed both for the fully data driven OTD equation but also the exact one, derived above (computed only for the purpose of numerical comparison).  In practice, the OTD system Eq.~\eqref{eq:otd-final} is integrated on a $64\times64$ spatial grid using a fourth-order Runge–Kutta (RK4) scheme.

\section{Results}
\label{sec:results}
We apply the proposed data-driven framework to the two-dimensional Kolmogorov flow using time-resolved snapshots of the velocity field $\{\mathbf{u}(t_k)\}_{k=0}^N$. Snapshots are obtained by integrating the Navier–Stokes equations~\eqref{eq:proj-NSE} in Fourier space over a time horizon of 30,000 time units, using a temporal discretization of $\Delta t = 0.1$ time units. This procedure yields a total of $N = 300{,}000$ state snapshots. Of these, 70\% are used for training the proposed framework, while the remaining 30\% are reserved for testing and performance evaluation. All results reported in this section correspond to the test dataset. Extreme events are quantified by the energy dissipation rate $D(t)$, which serves as the target observable, i.e. $z(t)=D(t)$.

We first examine the data-driven computation of FTLEs using the reduced-order formulation based on OTD modes, relying solely on the time-resolved snapshots of the velocity field.
 We then evaluate the predictive skill of the leading FTLE as a mechanism-based precursor by evaluating its ability to forecast the future evolution of $D(t)$ over the prescribed lead time $t+\tau$ with particular emphasis on its ability to capture extreme dissipation events.

\subsection{Reduced-order stability measures and precursors of dissipation}

The first step is to obtain a good approximation of the dynamics $\mathbf{\hat{F}}$. A detailed analysis of this step is provided in Appendix B.1. Based on the learned operator, we approximate the linearized operator $\mathscr{L}_{\mathrm{NS}}$ and proceed with the computation of the OTD modes. A comparison between the data-driven and equation-based OTD modes is presented in Appendix B.2.

\begin{figure}
    \centering
    \includegraphics[width=\linewidth]{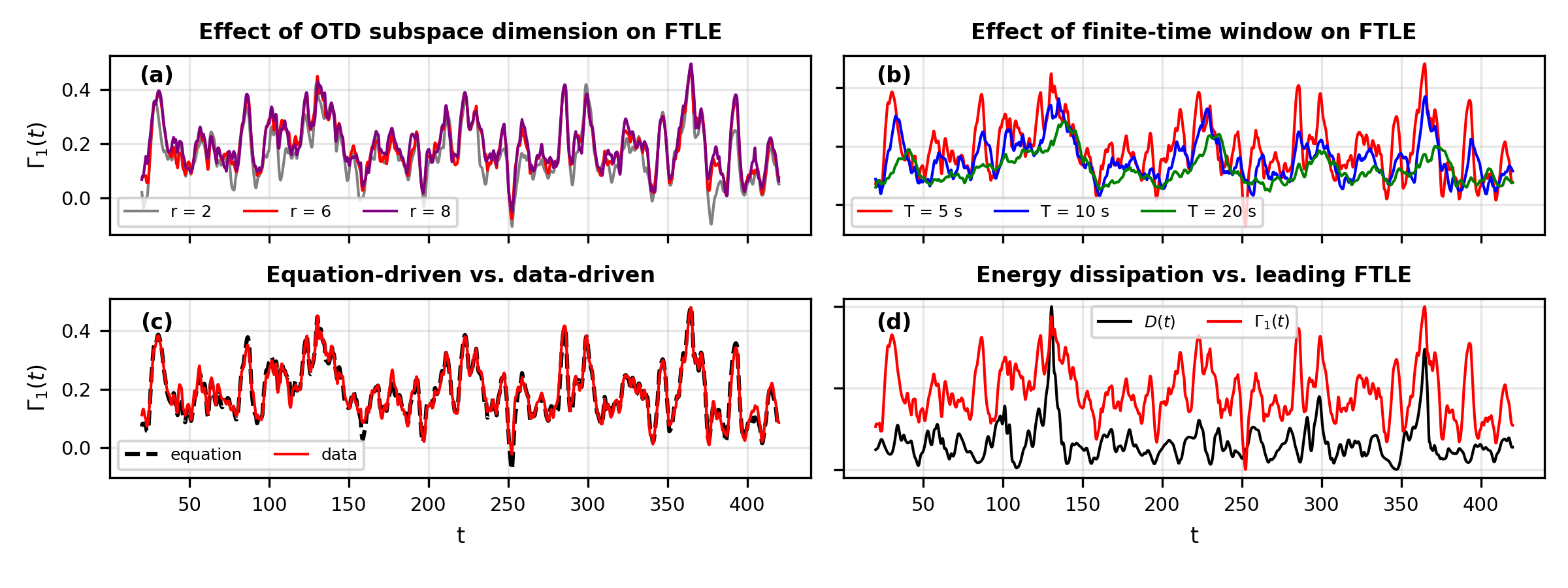}
    \caption{(a) Leading reduced-order finite-time Lyapunov exponent (FTLE), $\hat{\Gamma}_1$, computed within subspaces spanned by $r = 2,\,6,\,8$ OTD modes. The results show that the FTLE converges for $r \ge 6$, 
    whereas smaller subspaces ($r = 2$) underestimate transient growth rates. (b) Leading reduced-order FTLE, $\hat{\Gamma}_1$, computed over different integration horizons 
    $T = 5,\,10,$ and $20\,\mathrm{s}$; short horizons ($T = 5\,\mathrm{s}$) resolve a sequence of localized instability episodes that precede the main event, while increasing $T$ smooths these fluctuations, diminishing the FTLE’s 
    sensitivity as an early-warning indicator. (c) Comparison between data-driven computed FTLE and the one obtained from the analytical variational equations, for $r=6$. 
    (d) Superposition of the dominant FTLE and the observable of interest, $D(t)$ (both quantities are normalized for clarity). The close temporal alignment between the peaks demonstrates that the FTLE effectively captures the buildup of instability preceding dissipation bursts. 
}
    \label{fig:FTLE-comparison}
\end{figure}

\paragraph{Dependence of FTLE on the number of OTD modes and finite-time horizon, $T$.}
\label{subsec:results-ftle}
As outlined in the methodological framework of Section~\ref{sec:problem} (step~3), the reduced FTLEs are computed from the data-driven operator $\hat{\mathbf{L}}_r(t)$, which represents the projection of the learned tangent dynamics onto an $r$-dimensional OTD subspace. This reduced representation captures the most dynamically active directions responsible for transient growth and finite-time amplification. To evaluate how the dimensionality of the data-driven OTD subspace, $r$, influences the accuracy of the dominant FTLE, we consider three configurations with $r = 2, 6,$ and $8$ modes and evaluate the corresponding leading FTLE, $\hat{\Gamma}_1$, in each case. Figure~\ref{fig:FTLE-comparison}(a) displays these three cases. We note that the two-mode approximation underestimates the growth rate, indicating that it fails to encompass all dominant instability directions. Increasing the subspace dimension leads to rapid convergence of the estimated maximum FTLE, with the results for $r = 6$ and $r = 8$ being nearly identical. These findings indicate that a six-dimensional OTD subspace captures the essential transient growth dynamics with sufficient accuracy, representing the most efficient choice that balances fidelity and computational cost. A further comparison with the computed FTLE from the analytical OTD equations confirms that the fully data-driven accurately captures the OTD directions (Figure~\ref{fig:FTLE-comparison}(c)).

An additional quantity that is examined is the finite-time horizon $T$, over which the FTLEs are evaluated. To assess the sensitivity of the FTLEs to the interval length, the leading FTLE, $\hat{\Gamma}_1$, is computed for $T = 5$, 10, and 20~s. Figure~\ref{fig:FTLE-comparison}(b) shows these three representative cases. The shortest horizon ($T = 5$~s, red curve) yields the most distinct and temporally aligned precursor signal, effectively resolving the buildup of small-scale instabilities that precede dissipation bursts. Intermediate horizons ($T = 10$~s, blue curve) retain partial correspondence with the dissipation dynamics but exhibit increasing temporal smoothing and delay. The longest horizon ($T = 20$~s, green curve) performs the worst, producing a heavily smoothed and phase-lagged response that obscures short-term fluctuations. These results demonstrate that the predictive sharpness of $\hat{\Gamma}_1$ deteriorates with increasing $T$, emphasizing the need to select a horizon consistent with the characteristic time scales of transient amplification. Accordingly, $T = 5$~s is adopted in this work as the optimal horizon for subsequent analyses. 

Each FTLE represents the finite-time exponential growth rate of perturbations along a specific direction within the evolving OTD subspace. The leading FTLE, $\hat{\Gamma}_1$, corresponds to the most rapidly amplifying perturbation direction and thus reflects the dominant local instability mechanism at a given instant. Figure~\ref{fig:FTLE-comparison}(d) examines the temporal relationship between $\hat{\Gamma}_1$ and the energy dissipation rate, $D$. Peaks in $\hat{\Gamma}_1$ precede sharp increases in $D$, notably near $t \approx 120$~s and $t \approx 350$~s. This consistent lead–lag behavior reveals that transient instability episodes precede energy dissipation bursts, indicating the effectiveness of $\hat{\Gamma}_1$ as a precursor of upcoming extremes. From a physical perspective, the peaks in $\hat{\Gamma}_1$ correspond to intervals of intensified local stretching and strain amplification, during which perturbations grow rapidly and generate sharper velocity and vorticity gradients. The increase in gradient magnitude accelerates the transfer of energy toward smaller scales, thereby initiating the forward cascade that ultimately manifests as energy dissipation bursts.

\subsection{Energy dissipation prediction}

To assess the effectiveness of the selected precursor for long-horizon forecasting, we evaluate the Transformer-based model (Section~\ref{subsec:time-ser-for}) across multiple lead times $\tau \in \{2, 5, 7, 10, 12, 15\}$. The model input is a two-channel precursor sequence $\mathbf{\Pi}$, comprising the leading FTLE $\hat{\Gamma}_1$ and its time derivative $\hat{\Gamma}'_1$, sampled over a lookback window of length $\Delta$ up to the present time $t$. See Appendix A.2 for details of the hyperparameter tuning. The sequence-to-sequence model (Eq.~\eqref{eq:seq2seq}) predicts the evolution of energy dissipation $\hat{\mathbf{Z}}$ over the interval $[t,\,t+\tau]$; however, evaluation focuses on the terminal prediction $\hat{\mathbf{Z}}(t+\tau)$ as a measure of long-horizon predictive skill. Training over the full forecast window provides supervision at intermediate times, guiding the model to learn the continuous temporal evolution toward extreme events.

To evaluate the proposed FTLE-based precursor within the context of existing methodologies, we compare it against a Fourier-based precursor that has been widely used in prior studies. Specifically,~\cite{farazmand2017variational} identified the Fourier mode $\alpha(1,0)\in\mathbb{C}$ as an effective indicator of extreme events in Kolmogorov flow, with systematic reductions in $|\alpha(1,0)|$ preceding bursts in energy dissipation. This precursor has since been adopted in subsequent extreme-event forecasting studies. For instance,~\cite{asch2022model} utilized the real and imaginary components of $\alpha(1,0)$ as input channels in time-series models to predict extreme energy dissipation, demonstrating predictive capability over short forecast horizons ($\tau<5$). In this work, we compare this established Fourier-based approach with the proposed FTLE-based precursor to assess their relative performance at longer forecast horizons.

\begin{figure}
    \centering
    \includegraphics[width=0.7\linewidth]{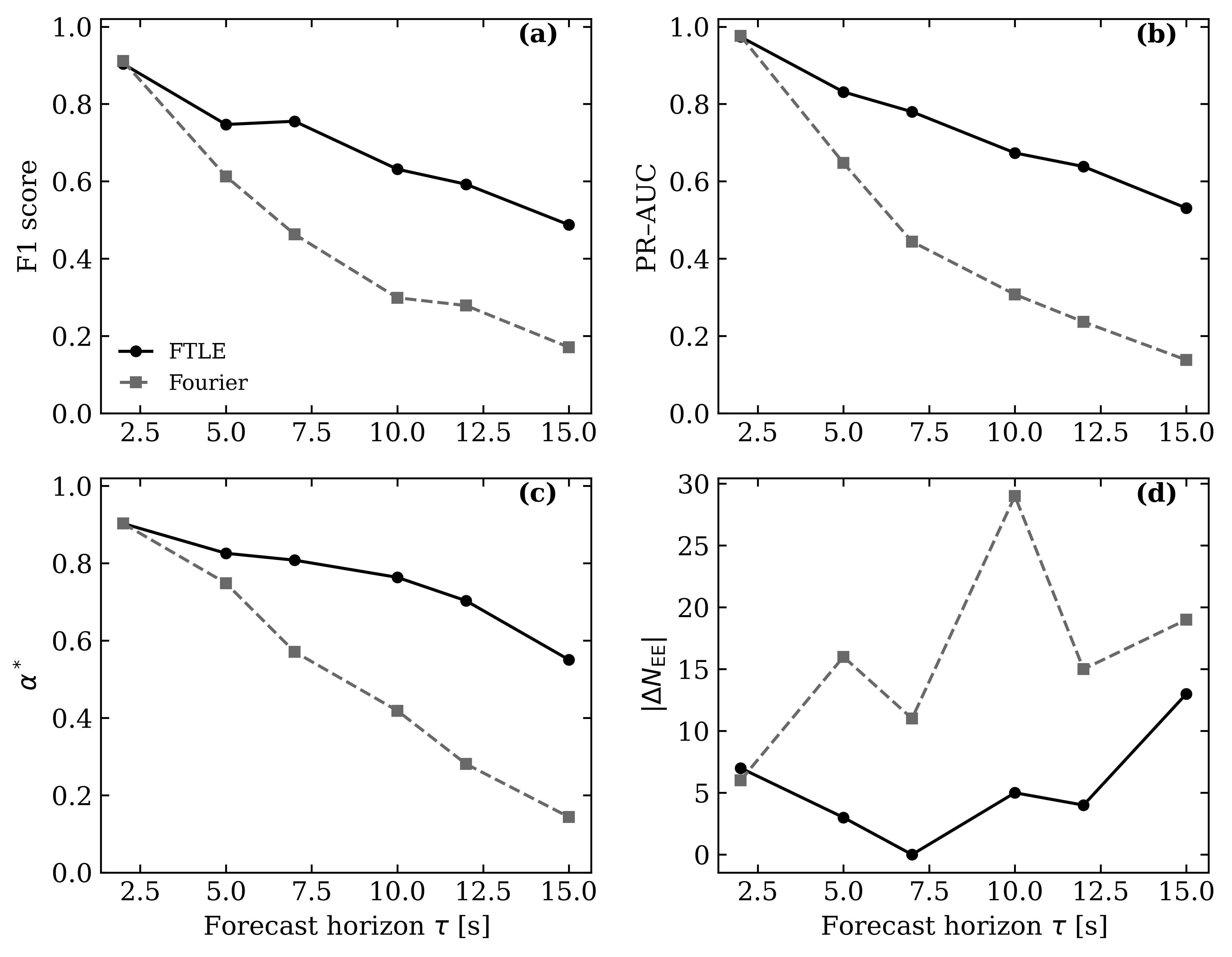}
    \caption{Performance comparison of binary classification metrics for forecasting extreme events using the FTLE-based precursor (black circles) and the Fourier-mode-based precursor (grey squares) as a function of the prediction horizon $\tau$. Panels show: (a) the F1-score, (b) the area under the precision–recall curve (AUC), (c) the adjusted AUC metric $\alpha^{*}$, and (d) the absolute deviation in the number of detected extremes between prediction and truth, $|\Delta N_{\mathrm{EE}}|$. The results indicate that the FTLE-based predictor maintains higher precision–recall performance and smaller event-count deviations across increasing forecast horizons, underscoring its superior robustness and predictive skill in capturing the onset of extreme events.}
    \label{fig:binry-crit}
\end{figure}

Extreme events are defined as instances in which the energy dissipation exceeds its mean by more than two standard deviations. To quantify extreme-event detection, we employ the binary classification metrics described in Section~\ref{sec:measures}; the F1-score, the area under the precision–recall curve (AUC), the adjusted AUC metric $\alpha^*$, and the absolute deviation in the number of detected extremes, $|\Delta N_{\mathrm{EE}}|$. Figure~\ref{fig:binry-crit} summarizes the results across prediction horizons $\tau$. All metrics exhibit decreasing performance with increasing lead time $\tau$, reflecting the growing difficulty of long-horizon prediction. Nevertheless, the FTLE-based predictor consistently outperforms the Fourier-mode-based model, achieving higher precision–recall performance and smaller event-count deviations across all horizons. This advantage is most pronounced for $\tau \leq 10$, where the FTLE-based approach maintains near-optimal F1 and AUC values. Although performance degrades for both methods at $\tau=15$, the FTLE-based predictor retains a substantial performance advantage.

\begin{figure}
    \centering
    \includegraphics[width=\linewidth]{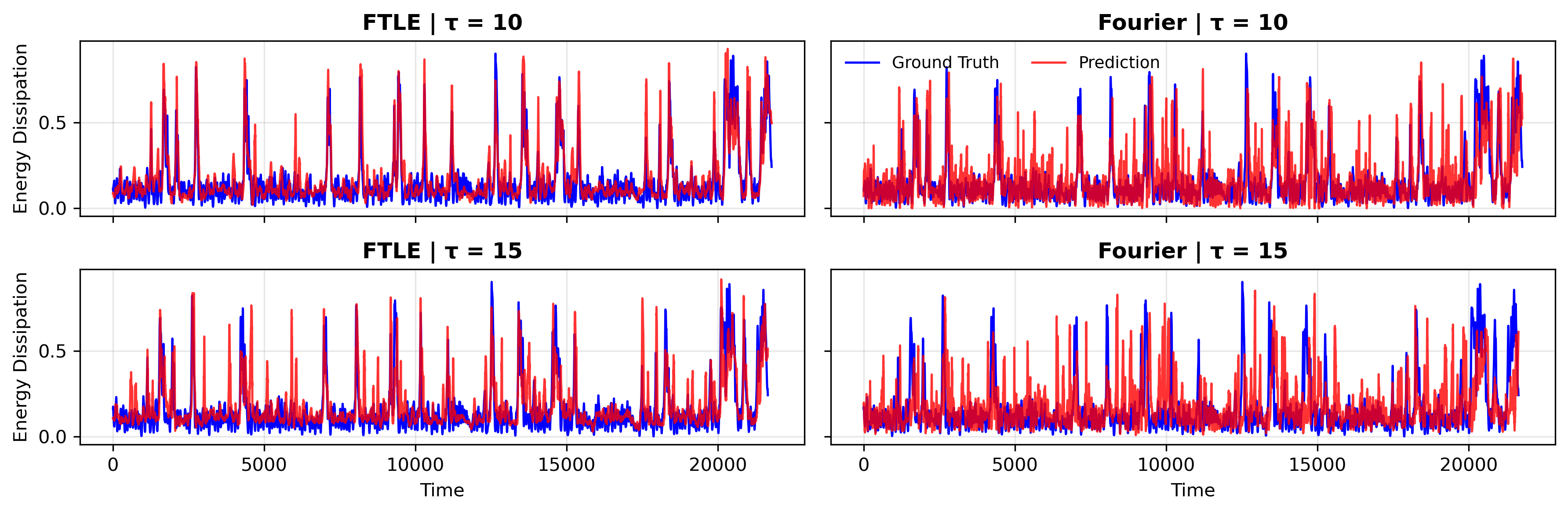}
    \caption{Predicted and true energy dissipation $D(t)$ at prediction horizons $\tau=10$ and $\tau=15$ time units. Blue color denote the ground-truth dissipation signal and red color indicate the model predictions. The left column shows forecasts obtained using the leading FTLE-based precursor while the right column corresponds to forecasts based on the Fourier coefficient approach~\cite{asch2022model}.}
    \label{fig:time-series}
\end{figure}

Figure~\ref{fig:time-series} compares predicted and true energy-dissipation signals for representative horizons $\tau=10$ and $\tau=15$. The left column shows forecasts based on the FTLE-based precursor, while the right column corresponds to the Fourier-mode precursor; blue curves denote ground truth and red curves the predictions. At $\tau=10$, the FTLE-based model accurately reproduces the dissipation dynamics, with predicted peaks closely aligned in timing, amplitude, and frequency with observed bursts. Even at $\tau=15$, the FTLE-informed forecasts remain largely coherent with the ground truth, capturing the timing and magnitude of most large-amplitude events, albeit with a modest increase in false positives. In contrast, forecasts based on the Fourier precursor deteriorate with increasing horizon: partial agreement is observed at $\tau=10$, while at $\tau=15$ temporal alignment is largely lost and fluctuations are overestimated, underscoring the limited ability of Fourier observables to constrain long-horizon dynamics

\begin{figure}
    \centering
    \includegraphics[width=0.7\linewidth]{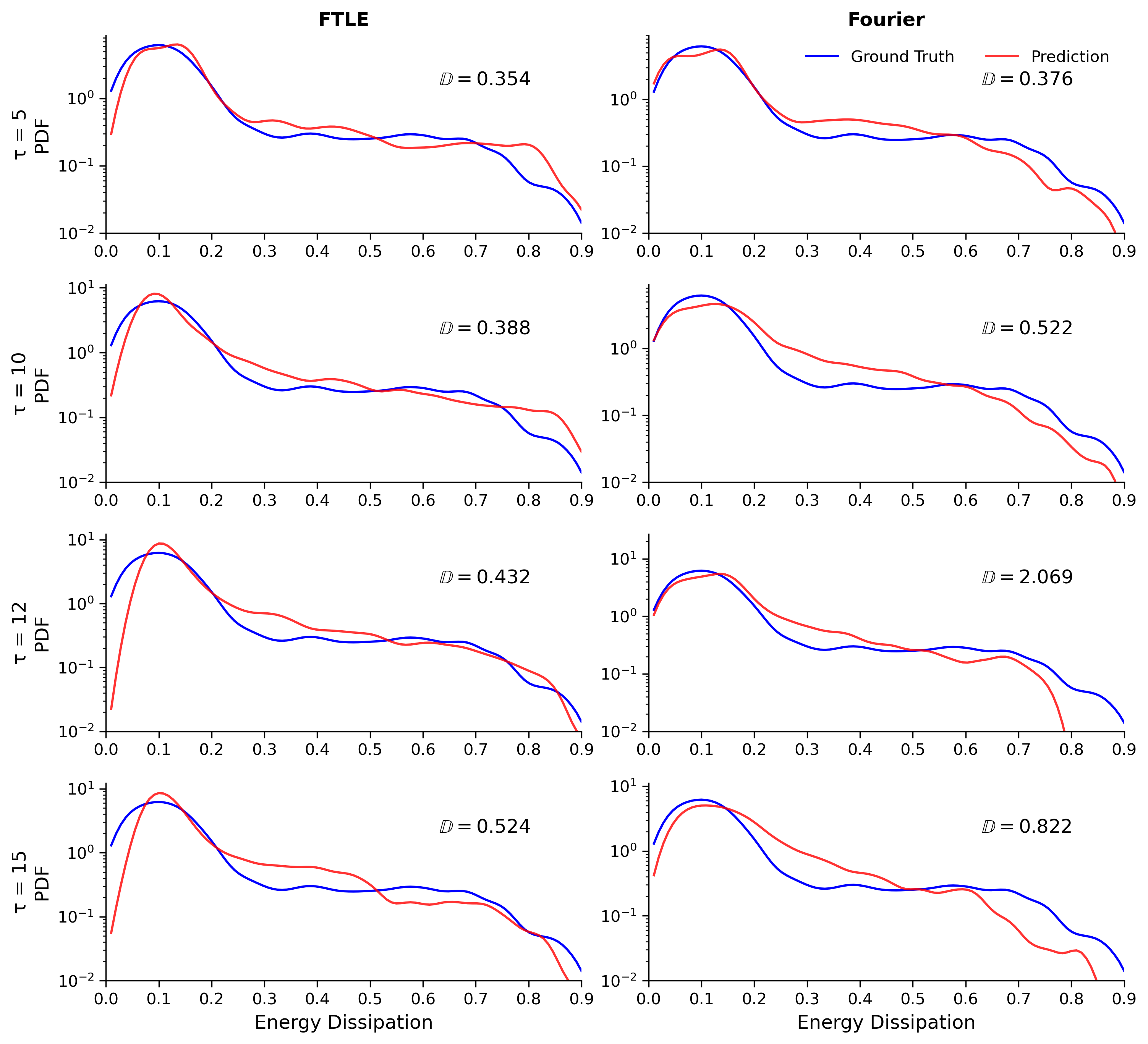}
    \caption{Probability density functions of the Kolmogorov flow energy dissipation $D(t)$ for different prediction horizons $\tau$. The left column shows results based on predictions using the leading FTLE precursor $\hat{\Gamma}_1(t)$, while the right column corresponds to predictions using the Fourier coefficient $\alpha(1,0,t)$. The blue curve represents the ground truth distributions, and the red curves corresponds to the predicted distributions. The reported value of $\mathbb{D}$ quantifies the discrepancy between the two PDFs and it is defined in Eq.~\eqref{eq:metric-D}.}
    \label{fig:PDF}
\end{figure}

Figure~\ref{fig:PDF} compares the probability density functions of the predicted and true energy dissipation for different lead times $\tau$, with discrepancies quantified by the metric $\mathbb{D}$ (Eq.~\eqref{eq:metric-D}). At short horizons ($\tau=5$), low $\mathbb{D}$ values indicate accurate reconstruction of both the bulk statistics and the onset of the tail. As $\tau$ increases, discrepancies arise primarily in the high-dissipation regime, where capturing heavy-tailed behavior becomes essential. The FTLE-based forecasts preserve good agreement in the tail up to $\tau \approx 10$–$12$, whereas the Fourier-based forecasts show pronounced tail attenuation and probability shifts toward moderate dissipation, resulting in a sharp increase in $\mathbb{D}$.



\section{Conclusions}
We have introduced a fully data-driven framework for long-horizon prediction of extreme events in high-dimensional chaotic dynamical systems, with emphasis on extremes generated by internal transient instabilities. The key idea is to move beyond purely statistical indicators and instead construct interpretable, dynamics-informed precursors that encode the dynamical pathways responsible for extreme-event formation.

We have relied on the concept of FTLEs which provide a natural description of the transient instability growth that precedes extreme events; however, since their classical computation is prohibitively expensive in high-dimensional systems, we adopt a reduced-order, data-driven approach based on OTD modes to compute FTLEs efficiently. Since the governing equations are unavailable and only state observations are assumed, the system dynamics required for OTD evolution are approximated directly from data and used solely to infer the local variational behavior. The resulting leading FTLE provides a physically grounded measure of finite-time instability growth, capturing the dynamical buildup preceding extreme excursions. We demonstrate how these FTLE-based precursors, computed using information available up to the present time, can be integrated with a Transformer-based architecture to enable long-horizon prediction of extreme events.

The developed architecture is applied to the two-dimensional Kolmogorov flow, where extremes manifest as intermittent bursts of energy dissipation. Across multiple evaluation criteria—including binary classification metrics and statistical consistency measures—the FTLE-based approach consistently achieves higher predictive skill and robustness, particularly at forecast horizons beyond those previously attainable with Fourier-based precursors. Specifically, when FTLE-based precursors are used, predictive performance degrades more gradually with increasing lead time. These results demonstrate that explicitly encoding transient instability mechanisms enables a meaningful extension of practical prediction horizons for rare extreme events. As the proposed framework relies only on time-resolved state observations and data-driven approximations of local dynamics, it is directly applicable to a broad class of high-dimensional systems in which extremes arise from transient instabilities, including turbulent flows, geophysical systems, and other complex multiscale dynamics. 

\section*{Funding Statement} The research has been supported by the Vannevar Bush Faculty Fellowship N000142512059 as well as the AFOSR grant FA9550-23-1-0517.

\section*{Data Accessibility} No external datasets were used. All results were generated computationally. The code required to reproduce the findings and figures is available at \url{https://github.com/Eirini-Katsidoniotaki/Precursors_Extreme_Events}.


\newpage
\section*{Appendix A: Details on the Methodology}

\subsection*{A.1 Approximating the system dynamics}
\label{app:approx-dynamics}
We construct a data-driven approximation of the nonlinear operator
\[
\mathbf{F} :\mathbf{u}(t) \mapsto \dot{\mathbf{u}}(t),
\]
which governs the instantaneous evolution of the system state as described in Eq.~\eqref{eq:setup_dynsystem}. Since the governing equations are not assumed to be available, $\mathbf{F}$ is inferred directly from observed state trajectories.

Let $\{\mathbf{u}(t_k)\}_{k=0}^N$ denote discrete snapshots of the system. Time derivatives are estimated from the data using a fourth-order central finite-difference scheme, yielding the training dataset
\[
    \mathcal{D} = \{(\mathbf{u}(t_k),\, \dot{\mathbf{u}}(t_k))\}_{k=1}^N.
\]
We note that $\mathbf{F}$ is an unbounded operator in the infinite-dimensional function space due to the presence of differential terms that amplify high-frequency components. This renders direct learning ill-conditioned in the continuous setting. To ensure numerical stability and well-posedness, the learning problem is formulated on the finite-dimensional state space $\mathcal{U}_d$ associated with the numerical discretization of the trajectory, where $\hat{\mathbf{F}}$ provides a stable approximation suitable for subsequent linearization and instability analysis.
We then learn an approximation of the dynamics via a nonlinear map
\[
\hat{\mathbf{F}} :\mathbf{u}(t) \mapsto \hat{\dot{\mathbf{u}}}(t),
\]

where $\hat{\mathbf{F}} :\mathcal{U}_d \to \mathcal{U}_d$. The learned operator $\hat{\mathbf{F}}$ is used to approximate the local linearized dynamics required for the computation of data-driven OTD modes. In particular, $\hat{\mathbf{F}}$ is substituted into the finite-difference formulation of Eq.~\eqref{eq:fin-diff} to evaluate Jacobian–vector products, $\mathbf{Lv}$, enabling the efficient computation of reduced-order instability measures.

Three neural architectures with complementary inductive biases are employed; their specific constructions are detailed in the following section. In the present study, the temporal resolution of the data is sufficiently fine to permit accurate estimation of time derivatives $\dot{\mathbf{u}}(t)$. For datasets sampled more coarsely in time, it may be preferable to infer the discrete flow map $\mathcal{S}: \mathbf{u}(t) \to \mathbf{u}(t+h)$ instead; see, e.g.,~\cite{tang2024learning, oommen2025learning, ovadia2025real}.


\subsection*{A.2 Time series forecasting}

\subsection*{A.2.1 Transformer-based model}
\label{subsec:transformer_precursor}

The objective is to learn a nonlinear operator that maps the temporal history \(\Delta\) of the precursor to the future evolution of the observable \(z(t)\) over a prediction horizon \(\tau\), discretized into \(n_\tau\) steps,
\[
\mathcal{N}_\theta:\;
\boldsymbol{\Pi}
\longmapsto
\hat{\mathbf{Z}}
=
\big[\, \hat{z}(t+\tfrac{\tau}{n_\tau}), \ldots, \hat{z}(t+\tau) \,\big]
\in \mathbb{R}^{n_\tau}.
\]

Each precursor vector is mapped to a latent representation via a learned convolutional embedding that captures local temporal structure, and augmented with a fixed sinusoidal positional encoding to preserve temporal ordering. We employ a time-series forecasting model based on the Transformer-based architecture~\cite{haoyietal-informer-2021, zhou2023expanding}. The model learns the operator~\(\mathcal{N}_\theta\) by constructing context-aware representations of the two-channel precursor sequence~\(\boldsymbol{\Pi}\). 

\paragraph{Encoder.}
The encoder embeds the precursor sequence into a latent space of dimension \(d\),
\[
\mathbf{E}^{(0)} = \boldsymbol{\Pi} W_{\mathrm{in}} + \mathbf{P},
\]
where \( W_{\mathrm{in}} \in \mathbb{R}^{2 \times d} \) is a learnable projection matrix and \( \mathbf{P} \) denotes positional encoding, which injects temporal ordering into the sequence.

Each encoder layer consists of a multi-head self-attention (MHSA) block followed by a position-wise feed-forward network (FFN), combined with residual connections and layer normalization:
\[
\begin{aligned}
\mathbf{Z}^{(\ell)} &= \mathbf{E}^{(\ell-1)} +
\mathrm{MHSA}\!\left(\mathbf{E}^{(\ell-1)}\right), \\
\mathbf{E}^{(\ell)} &= \mathbf{Z}^{(\ell)} +
\mathrm{FFN}\!\left(\mathbf{Z}^{(\ell)}\right).
\end{aligned}
\]

For each attention head, the queries, keys, and values are given by
\[
\mathbf{Q} = \mathbf{E} W_Q, \qquad
\mathbf{K} = \mathbf{E} W_K, \qquad
\mathbf{V} = \mathbf{E} W_V,
\]
and attention is computed as
\[
\mathrm{Attn}(\mathbf{Q}, \mathbf{K}, \mathbf{V})
=
\mathrm{softmax}\!\left(
\frac{\mathbf{Q}\mathbf{K}^\top}{\sqrt{d}}
\right)\mathbf{V}.
\]

The encoder output \( \mathbf{H}_{\mathrm{enc}} \) provides a compact representation of the precursor history, capturing both short- and long-range temporal dependencies in the instability dynamics.

\paragraph{Decoder.}

The decoder generates predictions conditioned on the encoded precursor representation. The initial decoder input sequence \( \mathbf{D}^{(0)} \) consists of the most recent \( n_\ell \) observed values of the target observable, followed by placeholders for the future prediction horizon, where \( n_\ell \) denotes the label length. Each decoder layer then applies masked self-attention, encoder–decoder (cross) attention, and a feed-forward network:
\[
\begin{aligned}
\mathbf{U}^{(\ell)} &= \mathbf{D}^{(\ell-1)} +
\mathrm{MaskedAttn}\!\left(\mathbf{D}^{(\ell-1)}\right), \\
\mathbf{V}^{(\ell)} &= \mathbf{U}^{(\ell)} +
\mathrm{CrossAttn}\!\left(\mathbf{U}^{(\ell)}, \mathbf{H}_{\mathrm{enc}}\right), \\
\mathbf{D}^{(\ell)} &= \mathbf{V}^{(\ell)} +
\mathrm{FFN}\!\left(\mathbf{V}^{(\ell)}\right).
\end{aligned}
\]

The masking enforces causality, ensuring that predictions at future times depend only on precursor information available up to the current forecast step. A final linear projection maps the decoder output to the predicted observable,
\[
\hat{\mathbf{Z}} = \mathbf{D}_{\mathrm{out}} W_{\mathrm{out}}.
\]

\paragraph{Informer-specific efficiency.}
To enable efficient learning from long precursor sequences, we adopt the Informer architecture, which replaces full self-attention with a probabilistic sparse attention mechanism. This reduces the computational complexity from \( \mathcal{O}(n_\Delta^2) \) to approximately \( \mathcal{O}(n_\Delta \log n_\Delta) \). In addition, convolutional distillation layers are employed within the encoder to progressively downsample the temporal dimension, retaining the most informative components of the precursor history while improving scalability.

\paragraph{Interpretation.}
From a dynamical systems perspective, the model learns a non-Markovian, data-driven approximation of the mapping
\[
\hat{z}(t+\tau)
\approx
\mathcal{N}_\theta\!\big(
\boldsymbol{\pi}(t), \boldsymbol{\pi}(t-\Delta t), \ldots
\big),
\]
where the attention mechanism adaptively identifies and weights the most dynamically informative segments of the precursor history for long-horizon forecasting of extreme events.

\subsection*{A.2.2 Setup}
\textbf{Look-back window length:} We performed a systematic study of the look-back length \(\Delta\) to evaluate its effect on predictive performance. Specifically, we tested look-back windows of $\Delta = 2\tau$, $3\tau$, $4\tau$, and $5\tau$, where $\tau$ denotes the characteristic time scale. The results showed that a look-back length of $4\tau$ provided the most accurate time series predictions, achieving the best performance under the AUC criterion. Throughout all experiments, the label length was fixed to \( n_\ell  = \frac{\Delta}{2}\).

\textbf{Hyperparameter tuning.}
In addition to the look-back window analysis, we performed hyperparameter tuning to optimize the model performance. The hyperparameters considered included the attention factor, embedding dimension, number of attention heads, numbers of encoder and decoder layers, feedforward network dimension, and dropout rate.
The attention factor was fixed to \(7\), which provides a favorable balance between contextual representation in the probabilistic attention mechanism and computational efficiency. The embedding dimension was varied in \(\{128, 256, 512\}\), while the number of attention heads was fixed at \(8\). The number of encoder layers was chosen from \(\{2,3,4\}\), and the number of decoder layers from \(\{2,3\}\). The feedforward network dimension was varied in \(\{256,512,1024\}\), and the dropout rate in \(\{0.05, 0.1, 0.2\}\).

Hyperparameter tuning was conducted using a random search strategy for prediction horizons \(\tau \in \{10,15\}\). Model performance was evaluated using the AUC metric. Based on predictive performance and computational efficiency, the final configuration was selected as follows: attention factor \(=7\), embedding dimension \(=256\), attention heads \(=8\), encoder layers \(=3\), decoder layers \(=3\), feedforward dimension \(=1024\), and dropout rate \(=0.1\). This configuration yielded stable training and consistently strong performance across prediction horizons.

\section*{Appendix B: Details on the Results}

\subsection*{B.1 Reconstruction of the approximated dynamics, $\hat{\mathbf{F}}$} 
\label{subsec:results-approx-dyn}

\subsubsection*{B.1.1 Deep learning models used for the approximation}
To reconstruct the data-driven operator $\hat{\mathbf{F}}$ of the Kolmogorov flow, we evaluate several neural architectures designed to infer the time derivative of the velocity field directly from state snapshots, $\hat{\dot{\mathbf{u}}} = \hat{\mathbf{F}}(\mathbf{u})$. The operator $\hat{\mathbf{F}}$ is learned on a $64 \times 64$ spectral grid defining the discrete state space $\mathcal{U}_d$, using velocity field snapshots spanning 30{,}000 time units with temporal resolution $\Delta t = 0.1$. Seventy percent of the data are used for training and the remaining thirty percent for testing, and all results reported in this section correspond to the test set.

All models are trained using the Adam optimizer with a batch size of 64, learning rate $10^{-3}$, and weight decay $10^{-4}$. A step-based learning-rate scheduler is applied to stabilize training by reducing the learning rate after a prescribed number of epochs. Training is carried out for 300 epochs, with periodic checkpointing to ensure reproducibility and limit overfitting.

\paragraph{Fourier Neural Operator (FNO).}
We employ the FNO~\cite{tang2024learning} to approximate the nonlinear operator
\[
\mathbf{F} : \mathcal{U}_d \rightarrow \mathcal{U}_d,
\qquad
\hat{\dot{\mathbf{u}}} = \hat{\mathbf{F}}(\mathbf{u}),
\]
owing to its ability to learn mappings between function spaces while preserving nonlocal and multiscale structure. In the present setting, the FNO provides a discrete realization of an operator-learning framework acting on the finite-dimensional subspace $\mathcal{U}_d$ induced by the spectral discretization.

Each FNO layer updates the feature field $\mathbf{u}^{(l)}$ according to
\[
\mathbf{u}^{(l+1)}(\mathbf{x})
=
\sigma\!\left(
W\,\mathbf{u}^{(l)}(\mathbf{x})
+
\mathcal{F}^{-1}\!\left(
R \cdot \mathcal{F}(\mathbf{u}^{(l)})
\right)(\mathbf{x})
\right),
\]
where $\mathcal{F}$ and $\mathcal{F}^{-1}$ denote the Fourier and inverse Fourier transforms, $R$ is a learnable linear operator acting on a truncated set of Fourier modes, $W$ is a pointwise linear map, and $\sigma$ is a nonlinear activation. This formulation enables efficient representation of global interactions through spectral convolutions while retaining local nonlinear effects via pointwise operations.

In our implementation, the FNO employs $n_{\mathrm{modes}}=(32,32)$ retained Fourier modes, $7$ Fourier layers, and $64$ hidden channels per layer, with two input and output channels corresponding to the velocity components. The model is trained using paired state–derivative samples, $\mathcal{D}$, to minimize the discrepancy between predicted and reference time derivatives.

\paragraph{Residual UNet++ (ResUNet++).} 
We employ the ResUNet++ architecture~\cite{jha2019resunet++}, an encoder--decoder convolutional network with residual and skip connections, to approximate the nonlinear operator $\hat{\mathbf{F}}$. The architecture incorporates residual blocks for stable training, squeeze-and-excitation (SE) modules for adaptive channel-wise feature recalibration, and attention mechanisms to enhance multiscale representation and focus on dynamically relevant regions.

Multiscale context is captured through an Atrous Spatial Pyramid Pooling (ASPP) module at the bottleneck, enabling aggregation of information across multiple receptive fields and supporting simultaneous representation of large-scale vortical structures and fine-scale gradients. The decoder employs attention-guided feature fusion and residual refinement to recover spatial detail while preserving global coherence. A final $1\times1$ convolution projects the decoded features to the predicted time derivative $\hat{\dot{\mathbf{u}}}$.

In our implementation, the network consists of four encoder stages with channel dimensions $(64,128,256,512)$ and three decoder stages with $(128,64,32)$. The ASPP module uses dilation rates $\{1,6,12,18\}$. All convolutions use $3\times3$ kernels with batch normalization and SiLU activations, replacing the original ReLU to improve smoothness and differentiability, which is beneficial for downstream Jacobian and OTD computations.

Overall, ResUNet++ provides a strong inductive bias for multiscale flow reconstruction, balancing global context and fine-scale spatial detail in the approximation of the Kolmogorov flow dynamics.

\paragraph{Residual CNN (ResCNN).}
As a baseline model, we employ a ResCNN to approximate the nonlinear operator $\hat{\mathbf{F}}$ using purely local spatial interactions. The architecture consists of an initial convolutional layer followed by six residual convolutional blocks, each containing two $3\times3$ convolutions with batch normalization, Tanh activation, and dropout. Residual skip connections enable the network to learn incremental corrections to the input representation, improving optimization stability and preserving spatial information across layers.

The network terminates with a final convolution projecting the features to the two output channels corresponding to the predicted time derivatives $\hat{\dot{\mathbf{u}}}_h$. While the ResCNN lacks explicit mechanisms to capture global or multiscale interactions, it provides a lightweight and computationally efficient baseline that isolates the role of local nonlinear feature extraction in modeling the Kolmogorov flow dynamics.

\paragraph{Loss function.} 
In chaotic PDE systems, like the Kolmogorv flow, the target field field exhibits strongly multiscale behavior, containing both low-frequency, large-scale structures and high-frequency, small-scale fluctuations. 
When training using the standard mean-square error or $L^2$ loss, the metric quantifies only the average difference in amplitude across space. Consequently, the model can reproduce the coarse, low-frequency patterns while neglecting fine-scale gradients or oscillations. However, these high-frequency details are precisely those that encode key physical quantities such as energy dissipation, vorticity, and turbulent structures. Minimizing only the $L^2$ loss therefore leads to over-smoothed predictions and a loss of physical accuracy. 

To address this limitation, several studies~\cite{czarnecki2017sobolev, beatson2020learning, li2021learning} have introduced Sobolev training, in which the loss function incorporates higher-order derivatives of the prediction error. This approach penalizes discrepancies not only in the field values but also in their spatial derivatives, thereby improving the representation of fine-scale physical features. We adopt the same principle in the training of our model.

For a function $f(x)$, the Sobolev norm of order $k$ with exponent $p$ is defined as
\[
\| f \|_{k,p} = \left( \sum_{i=0}^{k} \| f^{(i)} \|_{p}^{p} \right)^{1/p}
\]

For $p=2$, this becomes the Hilbert-space Sobolev norm, denoted by $H^k$,
\[
\| f \|_{H^k}^{2}
= \sum_{|\alpha| \le k} \| D^{\alpha} f \|_{L^{2}}^{2}.
\]
where the operator $D$ denotes a partial derivative of the function $f$ and $\alpha$ specifies the order of differentiation. In expanded form, this becomes
\[
\| f \|_{H^k}^{2}
= \| f \|_{L^{2}}^{2}
+ \| \nabla f \|_{L^{2}}^{2}
+ \| D^{2} f \|_{L^{2}}^{2}
+ \cdots
+ \| D^{k} f \|_{L^{2}}^{2}.
\]

In Fourier space, the same norm can be written as:
\[
\| f \|_{H^k}^{2} = \sum_{n=-\infty}^{\infty} \big( 1 + n^{2} + \cdots + n^{2k} \big) \, |\hat{f}(n)|^{2}.
\]
where $\hat{f}(n)$ denotes the Fourier coefficients of $f$. In this formulation, higher frequencies $n$ are weighted more heavily, since derivatives amplify the high-frequency content of the signal.

The order $k$ determines which spatial features are emphasized: $k=0$ corresponds to $L^2$ norm, which captures coarse-scale structures, $k=1$ (the $H^1$ norm) includes first derivatives, associated with velocity gradients, and $k=2$ (the $H^2$ norm) includes second derivatives, capturing  vorticity, and dissipation effects.

In our framework, we define 
\[
f = \hat{F}_h(\mathbf{u}_h) - F_h(\mathbf{u}_h)
\]
as the discrepancy between the predicted and reference velocity time derivatives. The training objective is then formulated as
\[
\mathcal{L}_{\text{Sob}} = \| f \|_{H^2}^2.
\]
This choice encourages the model to capture high-frequency information and higher-order spatial derivatives and moments, ensuring that it learns the correct small-scale flow structures, maintains accurate gradient information, and preserves the physical interpretability of the predicted dynamics.

\subsubsection*{B.1.2 Evaluate reconstruction accuracy} 
The accuracy of the learned map $\hat{\mathbf{F}}$ is assessed by comparing predicted time derivatives with their reference values on the test dataset. Three neural architectures are considered: a FNO, ResUNet++, and a ResCNN. Errors are quantified using Sobolev norms of order $k = 0, 1, \text{and}\, 2$, capturing discrepancies in field magnitude as well as first- and second-order spatial derivatives. 
\[
{H^k} =
\|\hat{\dot{\mathbf{u}}} - \dot{\mathbf{u}}\|_{H^k}
     , \qquad k = 0,1,2,
\]
These metrics provide a scale-aware measure of each model’s ability to reproduce both large-scale flow structures and fine-scale gradients. As reported in Table~\ref{tab:sobolev}, the FNO attains the lowest error across all Sobolev orders, indicating the most accurate approximation of the underlying dynamical operator. The ResUNet++ yields consistently higher, yet comparable, errors, while the ResCNN exhibits errors nearly an order of magnitude larger, reflecting limited capacity to represent the multiscale and nonlocal interactions inherent to the flow.

\begin{table}[h!]
\centering
\vspace{0.3cm}
\begin{tabular}{|l|c|c|c|}
\hline
\textbf{Model} & $\mathbf{H^0}$ & $\mathbf{H^1}$ & $\mathbf{H^2}$ \\ \hline
FNO         
& $\mathbf{2\times10^{-3}}$  
& $\mathbf{4\times10^{-3}}$  
& $\mathbf{1.4\times10^{-2}}$ \\ \hline

ResUNet++   
& $7\times10^{-3}$           
& $1\times10^{-2}$           
& $2.4\times10^{-2}$ \\ \hline

ResCNN      
& $7\times10^{-1}$           
& $9\times10^{-1}$           
& $1.3\times10^{0}$  \\ \hline
\end{tabular}

\vspace{0.2cm}
\caption{Comparison of learned operator accuracy measured in Sobolev norms 
$\| \hat{\dot{\mathbf{u}}} - \dot{\mathbf{u}} \|_{H^{k}}$ for $k=0,1,2$. 
}
\label{tab:sobolev}
\end{table}

\begin{figure}
    \centering
    \includegraphics[width=0.8\linewidth]{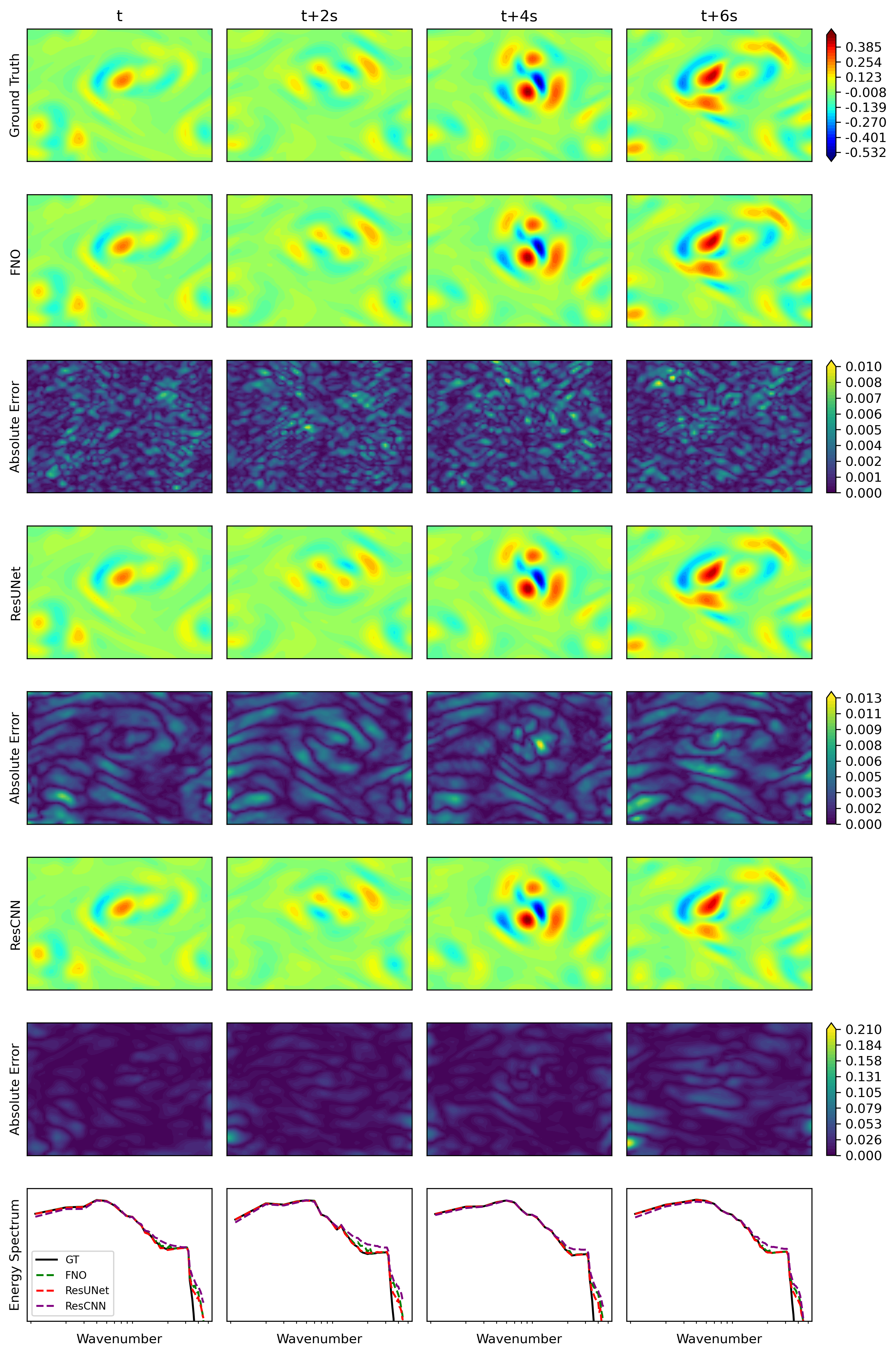}
    \caption{Comparison between the ground-truth and predicted time derivatives of the vorticity field, $\dot{\boldsymbol{\omega}} = \nabla \times \dot{\mathbf{u}}_h$, and $\hat{\dot{\boldsymbol{\omega}}} = \nabla \times \hat{\dot{\mathbf{u}}}_h$, obtained from the FNO, ResUNet++, and ResCNN surrogate models at $t$, $t{+}2\,\mathrm{s}$, $t{+}4\,\mathrm{s}$, and $t{+}6\,\mathrm{s}$. Each block of rows (top to bottom) shows the ground truth, model prediction, and absolute error, followed by the corresponding energy spectra comparing ground-truth and predicted fields.}
    \label{fig:curl_u_dot}
\end{figure}

Figure~\ref{fig:curl_u_dot} illustrates representative comparisons between reference and reconstructed time derivatives of the vorticity field, $\dot{\mathbf{\omega}} = \nabla \times \dot{\mathbf{u}}$ , at successive time instants. The vorticity derivative serves as a compact diagnostic of instantaneous dynamics, highlighting coherent vortical structures and sharp gradients associated with nonlinear interactions. The FNO and ResUNet++ accurately reproduce the dominant structures and their temporal evolution, whereas the ResCNN systematically underestimates local gradients, resulting in smoother fields and increased localized error. Absolute-error maps remain approximately an order of magnitude smaller than the signal amplitude, confirming faithful recovery of the dominant flow features.

The corresponding energy spectra (bottom row of Fig.~\ref{fig:curl_u_dot}) provide a complementary, scale-wise assessment. The spectra obtained from the FNO and ResUNet++ closely match the reference solution over a broad range of wavenumbers, with deviations confined to the highest, dissipative scales. These discrepancies reflect the inherent smoothness bias of neural architectures, arising from spectral truncation, convolutional filtering, and smooth activation functions. In the present configuration, the Kolmogorov flow is dominated by low-frequency modes, and all models adequately resolve the dynamically relevant scales. For flows exhibiting stronger high-frequency activity or slower spectral decay, architectures specifically designed to enhance fine-scale spectral fidelity may be required~\cite{oommen2025learning}.

The results indicate that $\hat{\mathbf{F}}$ provides an accurate, stable, and physically consistent approximation of the true dynamics. This fidelity enables reliable evaluation of the Jacobian–vector products $\mathbf{Lv}$, which serve as the foundation for the computation of the OTD modes.

\subsection*{B.2 Data-driven OTD modes} 
\begin{figure}
    \centering
    \includegraphics[width=0.3\linewidth]{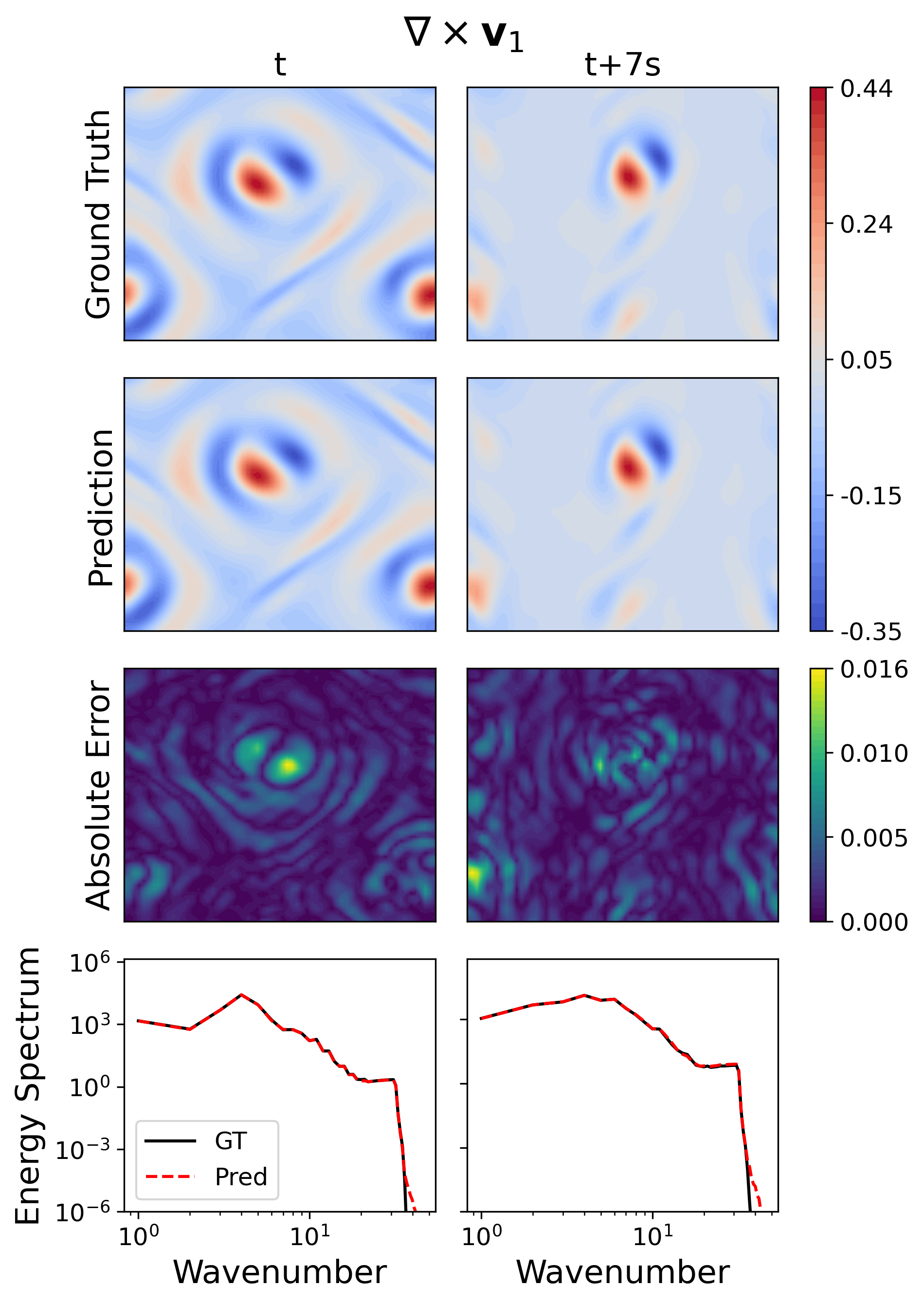}
    \includegraphics[width=0.3\linewidth]{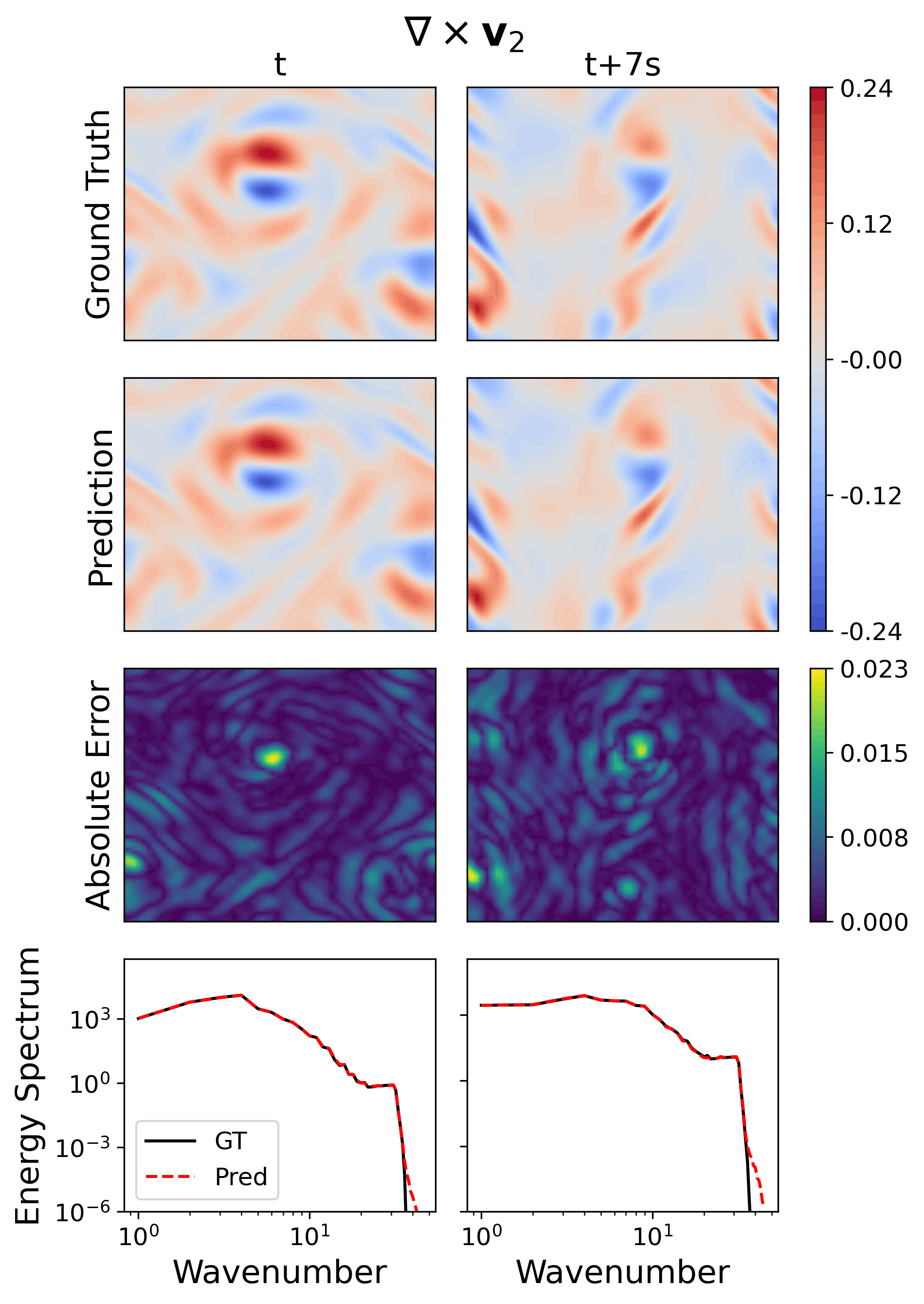}
    \includegraphics[width=0.3\linewidth]{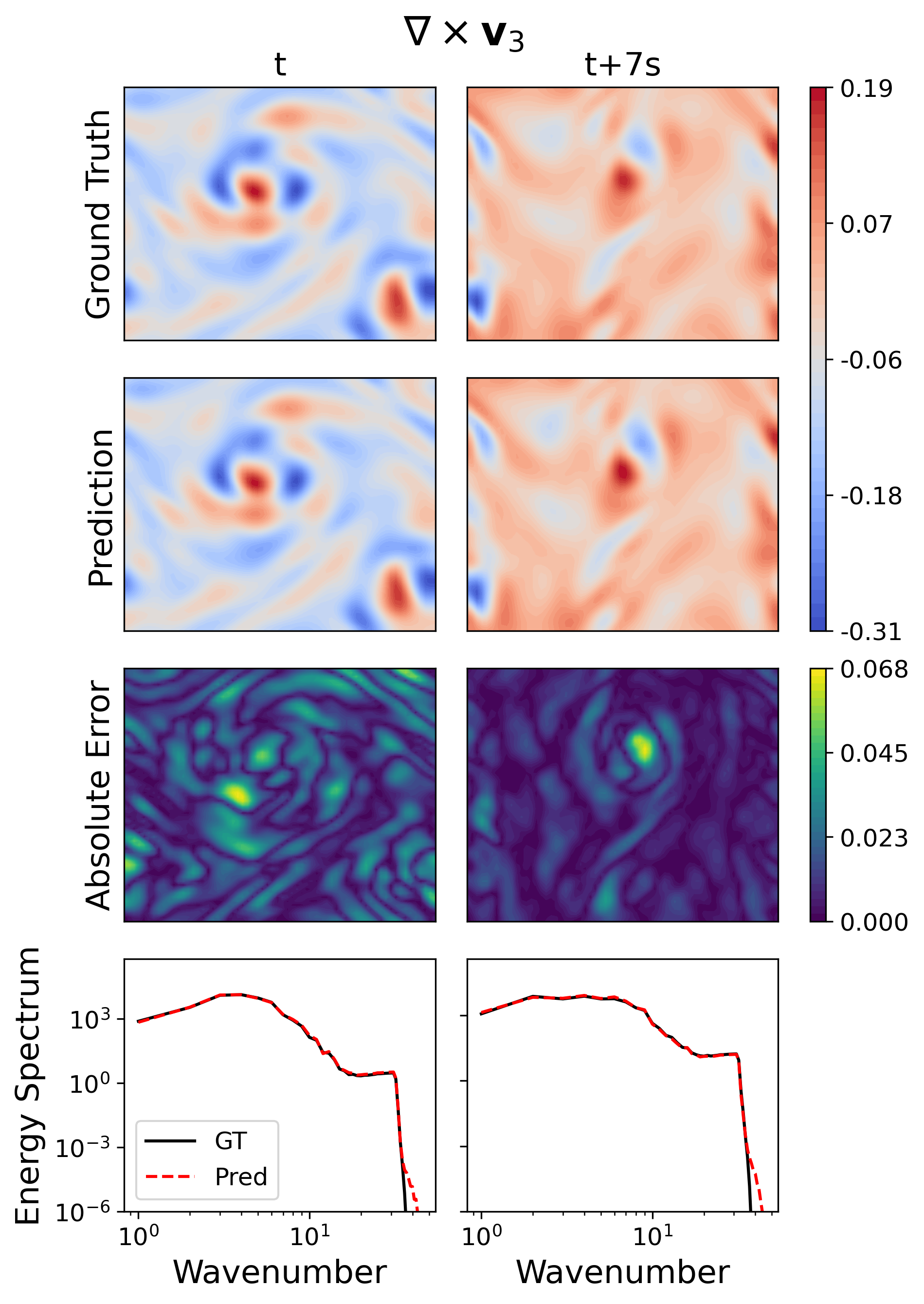}
    \includegraphics[width=0.3\linewidth]{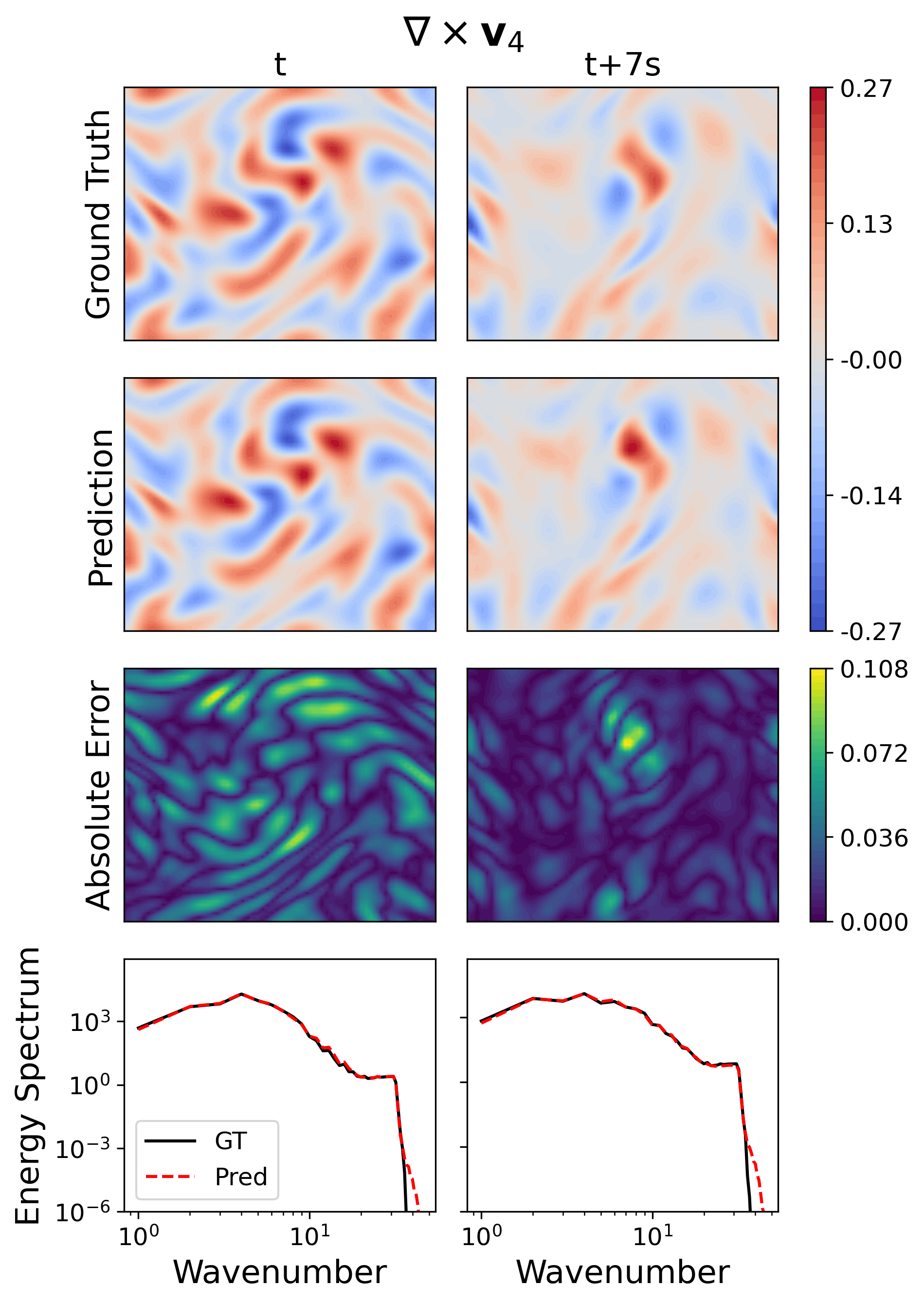}
    \includegraphics[width=0.3\linewidth]{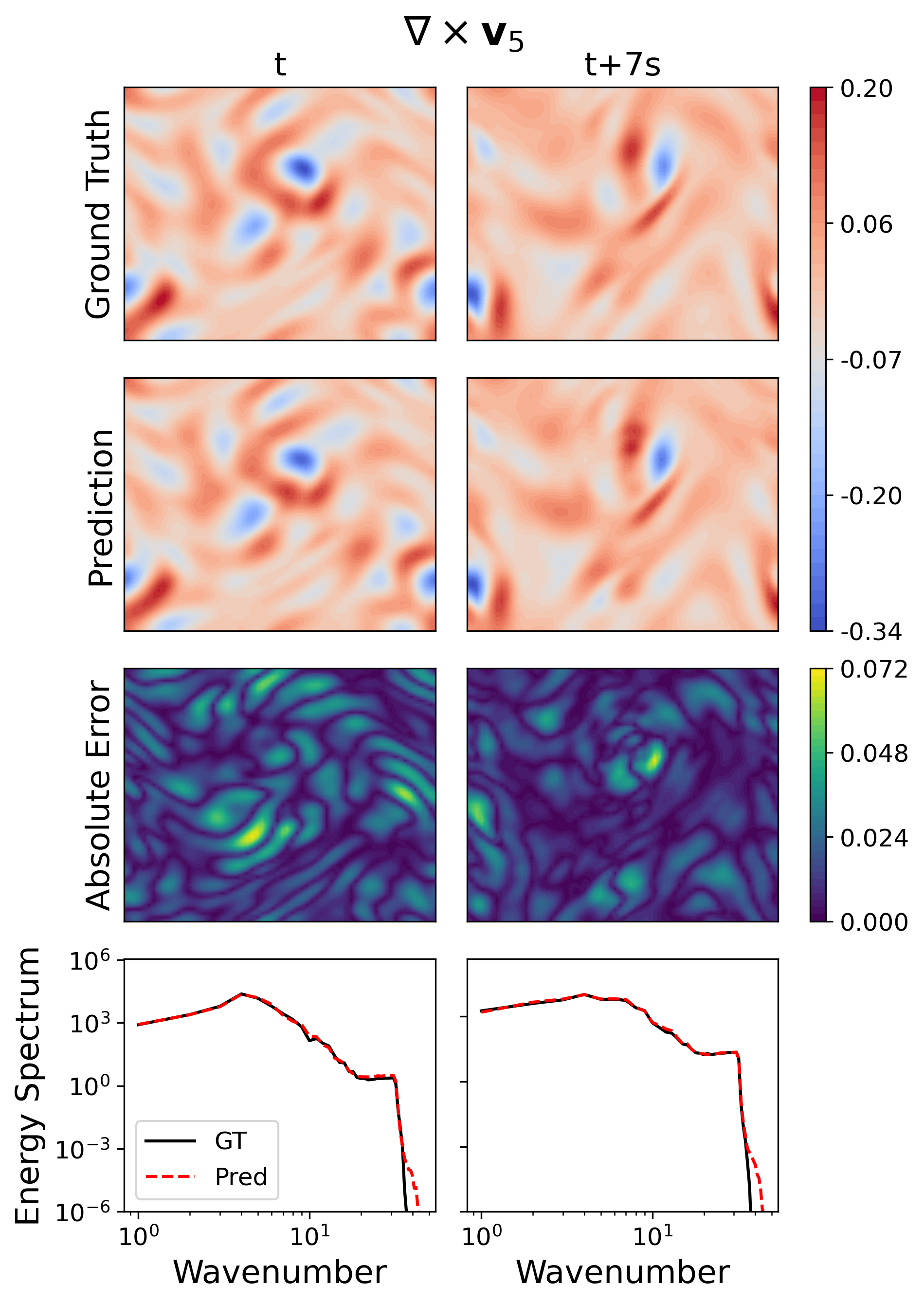}
    \includegraphics[width=0.3\linewidth]{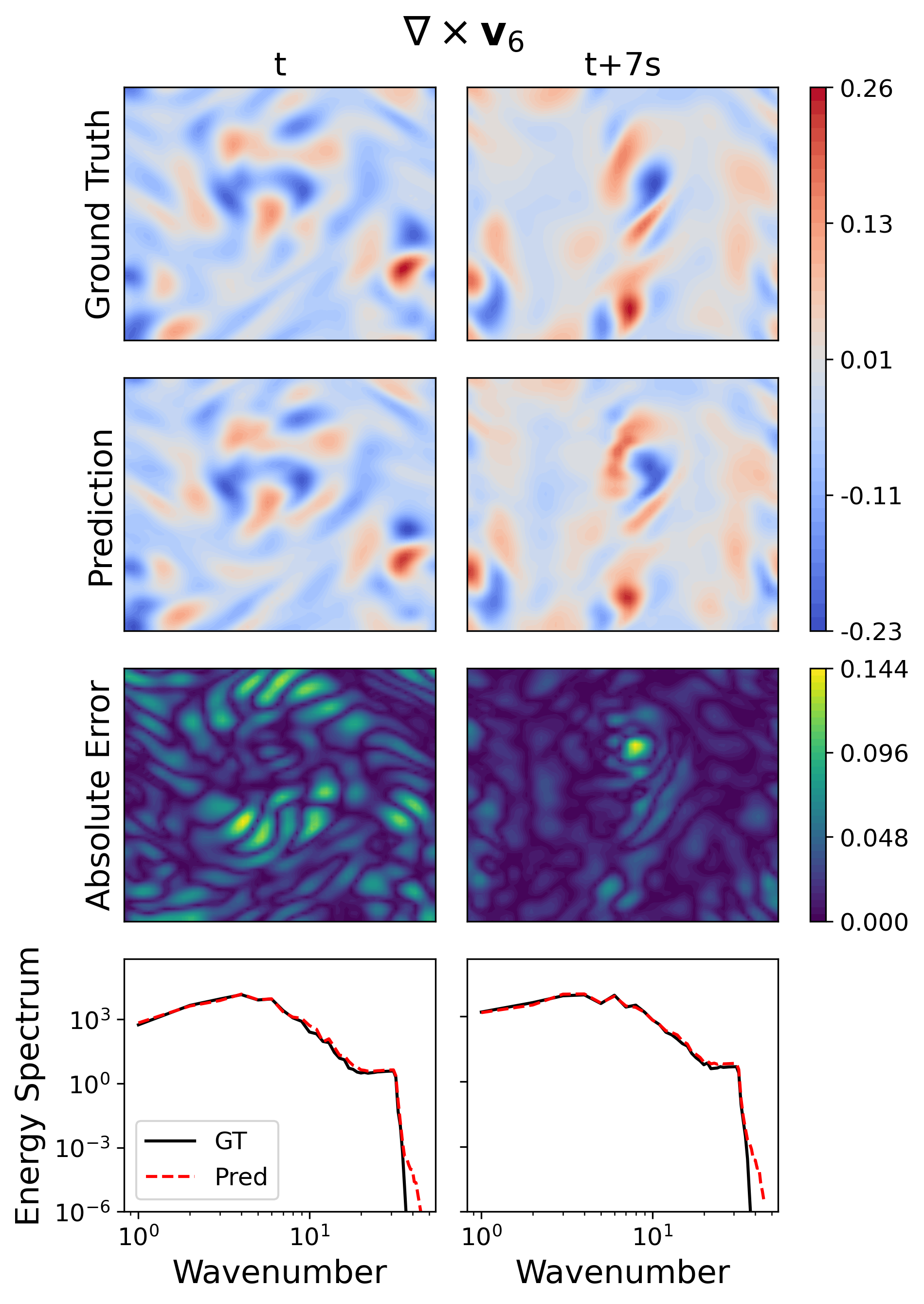}
    \caption{Evolution of six OTD modes over a 7-second horizon, visualized through 
    the curl of the mode fields, $\nabla \times \mathbf{v}_i(t)$, for $i=1,\dots,6$. 
    Each column corresponds to successive time instants, while rows show (top to bottom): 
    ground truth, predictions obtained from the data-driven OTD framework using the 
    FNO-based surrogate operator, and the absolute error. The bottom panels present 
    the corresponding energy spectra, comparing predicted and ground-truth modes.}
    \label{fig:curl_otd}
\end{figure}

Figure~\ref{fig:curl_otd} compares the curl of the first six OTD modes of the Kolmogorov flow,
$\nabla\times\mathbf{v}_i(t)$ for $i=1,\ldots,6$, at two instants.
For each mode, the top two rows show the equation-based and the data-driven prediction; the third row shows the absolute error; the bottom row reports the energy spectra at each time.

The leading OTD modes, $\mathbf{v}_1$ and $\mathbf{v}_2$, represent the most energetic directions of instantaneous instability growth and are characterized by large, coherent vortical structures. The data-driven predictions accurately reproduce their spatial organization and temporal evolution between $t$ and $t+7$, preserving orientation and circulation strength. Errors are weak and localized near vortex peripheries, consistent with small phase offsets rather than amplitude discrepancies, and the predicted and reference energy spectra show near-perfect agreement across the resolved wavenumber range.

Higher-order modes ($\mathbf{v}_3$–$\mathbf{v}_6$) exhibit increasingly intricate and less coherent flow patterns associated with weaker instability directions. While localized discrepancies become more pronounced with increasing mode index—particularly near regions of strong gradients and filamentary structures—the data-driven modes retain the correct overall structure and temporal evolution. The corresponding energy spectra remain in close agreement with the reference solution across the energy-containing and inertial ranges, indicating that the essential multiscale structure of the instability subspace is preserved.

\bibliographystyle{plain}
\bibliography{references}
\end{document}